%% file: Main.tex
\documentclass[sigconf]{acmart}
\renewcommand\footnotetextcopyrightpermission[1]{} 
\usepackage{nopageno}

\usepackage{booktabs} 
\usepackage{enumerate}
\usepackage{amsmath}
\usepackage{amsfonts}
\usepackage{graphicx}
\usepackage[ruled,vlined]{algorithm2e}
\usepackage{algpseudocode}
\usepackage{bm}
\usepackage{silence}
\usepackage{subfigure}
\usepackage{multirow}
\usepackage{enumitem}
\usepackage{array}
\usepackage{xcolor}
\usepackage{balance}
\newtheorem{problem}{Problem Definition}

\makeatletter
\renewcommand{\ALG@beginalgorithmic}{\small}
\makeatother

\newcommand{\squishlist}{
 \begin{list}{$\bullet$}
  { \setlength{\itemsep}{0pt}
     \setlength{\parsep}{3pt}
     \setlength{\topsep}{3pt}
     \setlength{\partopsep}{0pt}
     \setlength{\leftmargin}{1.5em}
     \setlength{\labelwidth}{1em}
     \setlength{\labelsep}{0.5em} } }

\newcommand{\squishlisttwo}{
 \begin{list}{$\bullet$}
  { \setlength{\itemsep}{0pt}
     \setlength{\parsep}{0pt}
    \setlength{\topsep}{0pt}
    \setlength{\partopsep}{0pt}
\setlength{\leftmargin}{2em}
\setlength{\labelwidth}{1.5em}
\setlength{\labelsep}{0.5em} } }

\newcommand{\squishend}{
\end{list}  }

\copyrightyear{2020} 
\acmYear{2020} 
\setcopyright{acmcopyright}\acmConference[CIKM '20]{Proceedings of the 29th ACM International Conference on Information and Knowledge Management}{October 19--23, 2020}{Virtual Event, Ireland}
\acmBooktitle{Proceedings of the 29th ACM International Conference on Information and Knowledge Management (CIKM '20), October 19--23, 2020, Virtual Event, Ireland}
\acmPrice{15.00}
\acmDOI{10.1145/3340531.3411922}
\acmISBN{978-1-4503-6859-9/20/10}

\begin{document}
\title{Graph Prototypical Networks for Few-shot Learning\\ on Attributed Networks}

\author{Kaize Ding}
\affiliation{%
  \institution{Arizona State University}
}
\email{kaize.ding@asu.edu}

\author{Jianling Wang}
\affiliation{%
  \institution{Texas A\&M University}
}
\email{jlwang@tamu.edu}

\author{Jundong Li}
\affiliation{%
  \institution{University of Virginia}
  }
\email{jundong@virginia.edu}

\author{Kai Shu}
\affiliation{%
  \institution{Illinois Institute of Technology}
  }
\email{kshu@iit.edu}

\author{Chenghao Liu}
\affiliation{%
  \institution{Singapore Management University}
  }
\email{chliu@smu.edu.sg}

\author{Huan Liu}
\affiliation{%
  \institution{Arizona State University}
  }
\email{huan.liu@asu.edu}


%
%

\begin{abstract}
    \input{Abstract}
\end{abstract}


\maketitle

\section{Introduction}

\input{Intro}

\section{Related Work}
\input{Related.tex}
 
\section{Problem Statement}
\input{Problem.tex}

\section{Graph Prototypical Networks}

\input{Methodology.tex}

\section{Experiments}
\input{Experiments.tex}

\section{Conclusion}
\input{Conclusion.tex}



\balance
\bibliographystyle{acm}
\bibliography{acmart}

\appendix
\input{Appendix.tex}

\end{document}

%% file: Abstract.tex
Attributed networks nowadays are ubiquitous in a myriad of high-impact applications, such as social network analysis, financial fraud detection, and drug discovery. As a central analytical task on attributed networks, node classification has received much attention in the research community. In real-world attributed networks, a large portion of node classes only contains limited labeled instances, rendering a long-tail node class distribution. Existing node classification algorithms are unequipped to handle the \textit{few-shot} node classes. As a remedy, few-shot learning has attracted a surge of attention in the research community. Yet, few-shot node classification remains a challenging problem as we need to address the following questions: (i) How to extract meta-knowledge from an attributed network for few-shot node classification? (ii) How to identify the informativeness of each labeled instance for building a robust and effective model? To answer these questions, in this paper, we propose a graph meta-learning framework -- Graph Prototypical Networks (GPN). By constructing a pool of semi-supervised node classification tasks to mimic the real test environment, GPN is able to perform \textit{meta-learning} on an attributed network and derive a highly generalizable model for handling the target classification task. Extensive experiments demonstrate the superior capability of GPN in few-shot node classification.


%% file: Intro.tex
Due to its strong modeling capability, attributed networks have been increasingly used to model a myriad of graph-based
systems, such as social media networks~\cite{qi2011exploring,ding2019interactive}, citation networks~\cite{tang2008arnetminer,ding2020inductive}
and gene regulatory networks~\cite{subramanian2005gene}. Among various analytical tasks on attributed networks, node classification is an essential one that has a broad spectrum of applications, including social circle learning~\cite{leskovec2012learning}, document categorization~\cite{tang2008arnetminer}, and protein classification~\cite{borgwardt2005protein}, to name a few. Briefly, the objective is to infer the missing labels of nodes given a partially labeled attributed network. To tackle this problem, plenty of approaches have been proposed in the research community and demonstrated promising performance~\cite{perozzi2014deepwalk,grover2016node2vec,kipf2016semi,velickovic2017graph}. 

\begin{figure}[t]
    \graphicspath{{figures/}}
    \centering
    \includegraphics[width=0.875\linewidth]{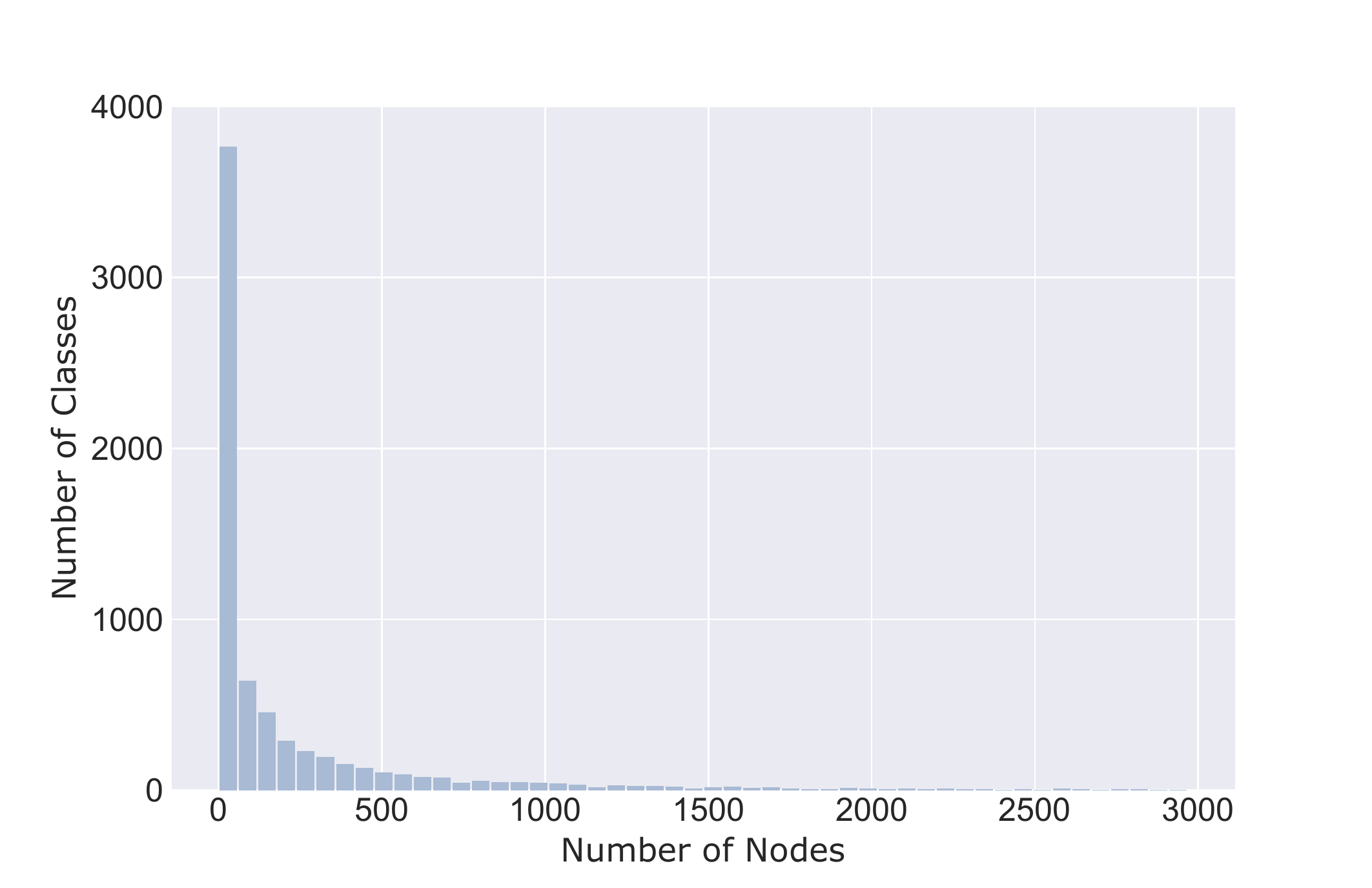}
    \caption{The histogram shows the distribution of labeled nodes in the real-world DBLP dataset.}%
    \label{fig:long-tail}%
    \vspace{-0.3cm}
\end{figure}


Prevailing approaches for the node classification problem usually follow a supervised or semi-supervised paradigm, which typically relies upon the availability of sufficient labeled nodes for all the node classes~\cite{zhou2019meta}. Nonetheless, in many real-world attributed networks, a large portion of node classes only contain a limited number of labeled instances, rendering a long-tail distribution of node class labels. As shown in Figure \ref{fig:long-tail}, DBLP~\cite{tang2008arnetminer} is a dataset where nodes represent publications and node labels denote venues. Among all the node classes, more than $30\%$ of them have less than $10$ labeled instances. In the meantime, many practical applications require the learning models to possess the capability of dealing with such \textit{few-shot} classes. A typical example is the intrusion detection problem~\cite{northcutt2002network,garcia2009anomaly} on traffic networks, where new attacks and threats are continuously being developed by adversaries. Due to the intensive labeling cost, for a specific type of attack, only a few examples can be accessed. Thus, understanding those attacks by type with limited labeled data is crucial for providing effective countermeasures. The shortage of labeled training data hinders existing node classification algorithms from learning an effective model with those \textit{few-shot} node classes~\cite{zhou2019meta,qiao2019transductive,yao2019graph}. As such, it is challenging yet imperative to investigate the problem of node classification on attributed networks under the \textit{few-shot} setting.

Recently, much research progress has been made in few-shot learning (FSL), for solving tasks (e.g., classification) with only a handful of labeled examples. In general, an FSL model learns across diverse \textit{meta-training} tasks sampled from those classes with a large quantity of labeled data, and can be naturally generalized to a new task (i.e., \textit{meta-test} task) from unseen classes during training. Such a \textit{meta-learning} procedure enables the model to adapt knowledge from previous experiences, and has led to significant progress in FSL problems. Specifically, a major line of research such as siamese networks~\cite{koch2015siamese}, matching networks~\cite{vinyals2016matching}, and relation networks~\cite{sung2018learning} attempts to make the prediction by comparing the query instances and labeled examples in a shared metric space. These \textit{learning-to-compare} approaches have come into fashion due to their simplicity and effectiveness. 


Despite their fruitful success, few-shot learning on attributed networks remains largely unexplored, mainly because of the following two challenges: \textbf{(i)} The process of constructing those \textit{meta-training} tasks depends on the assumption that data is independent and identically distributed (\textit{i.i.d.}), which is invalid on attributed networks. Apart from conventional text or image data, attributed networks lie in non-Euclidean space and encode the inherent dependency between nodes. Directly grafting existing methods is infeasible to capture the underlying data structure, making the embedded node representations less expressive. Thus how to exert the power of \textit{meta-learning} on attributed networks is indispensable for extracting the meta-knowledge from data; \textbf{(ii)} Most of the existing FSL approaches simply assume that all the labeled examples are of equal importance for characterizing their belonged classes. However, neglecting the individual informativeness of labeled nodes will inevitably restrict the model performance on real-world attributed networks: On the one hand, it makes the FSL model highly vulnerable to noises or outliers since labeled data is severely limited~\cite{ren2018meta,zhang2019variational}; on the other hand, it runs counter to the fact that the significance of a node could largely deviate from another. Intuitively, those central (core) nodes in a community are supposed to be more representative~\cite{zhang2012method}. Hence, how to capture the informativeness of each labeled node is the other challenge for building an effective few-shot classification model on attributed networks.

To address the aforementioned challenges, we present Graph Prototypical Networks (GPN), a graph meta-learning framework for solving the problem of few-shot node classification on attributed networks. Instead of classifying nodes directly, GPN tries to learn a transferable metric space in which the label of a node is predicted by finding the nearest class prototype. The proposed framework consists of two essential components that seamlessly work together for learning the prototype representation of each class. Specifically, the \textit{network encoder} in GPN first compresses the input network to expressive node representations via graph neural networks (GNNs), in order to capture the data heterogeneity of an attributed network. Concurrently, another GNN-based \textit{node valuator} is developed to estimate the informativeness of each labeled instance, by leveraging additional information encoded in the network. 
In this way, GPN derives highly robust and representative class prototypes. Moreover, by performing \textit{meta-learning} across a pool of semi-supervised node classification tasks, GPN gradually extracts the meta-knowledge on an attributed network and further achieves better generalization ability on the target few-shot classification task. In summary, the main contributions of our work are as follows:





\begin{itemize}[leftmargin=*,noitemsep,topsep=1.5pt]


\item \textbf{\emph{Problem}}: We investigate the novel problem of few-shot node classification on attributed networks. In particular, we emphasize its importance in real-world applications and further provide a formal problem definition.

\item \textbf{\emph{Algorithm}}: We propose a principled framework GPN for the problem, which exploits graph neural networks and meta-learning to learn a powerful few-shot node classification model on attributed networks.


\item \textbf{\emph{Evaluation}}: We perform extensive experiments on various real-world datasets to corroborate the effectiveness of our approach. The experimental results demonstrate the superior performance of GPN for few-shot node classification on attributed networks.

\end{itemize}

%% file: Related.tex
In this section, we briefly summarize related work into two categories: (1) graph neural networks; and (2) few-shot learning.

\subsection{Graph Neural Networks}
Driven by the momentous success of deep learning, recently, a mass of efforts have been devoted to developing deep neural networks for graph-structured data~\cite{chang2015heterogeneous,cao2016deep,ding2019deep,wang2020next}. As one of the pioneer works, GNN~\cite{scarselli2009graph} was introduced to learn node representations by propagating neighbors' information via recurrent neural architecture. Based on the graph spectral theory, a series of graph convolutional networks (GCNs) have emerged and demonstrated superior learning performance by designing different graph convolutional layers. Among them, the first prominent research on GCNs called Spectral CNN~\cite{bruna2013spectral}, which extends
the convolution operation in the spectral domain for network representation learning. Since then, increasing research advances on graph convolutional networks~\cite{kipf2016semi,defferrard2016convolutional,henaff2015deep} are presented as its extensions. In addition to spectral graph convolution models, graph neural networks that follow neighborhood aggregation schemes are also extensively investigated. Instead of training individual embeddings for each node,
those methods learn a set of \textit{aggregator functions} to aggregate features from a node's local neighborhood. GraphSAGE~\cite{hamilton2017inductive} learns a function that generates embeddings by sampling and aggregating features
from a node's local neighborhood. Similarly, Graph Attention Networks (GATs)~\cite{velickovic2017graph} incorporate trainable attention weights to specify fine-grained weights on neighbors when aggregating neighborhood information of a node. Furthermore, Graph Isomorphism Network (GIN)~\cite{xu2018powerful} extends this idea with arbitrary aggregation functions on multi-sets, and is proven to be as theoretically powerful as the  Weisfeiler-Lehman (WL) graph isomorphism test. Nevertheless, all the existing GNN models focus on semi-supervised node classification. The inability to handle unseen classes with severely limited samples, is one of the major challenges for the current GNNs. In this paper, we propose a novel GNN framework to tackle the problem of few-shot node classification on graph-structured data.

\begin{table}
  
  \label{tab:symbols}
\caption{Table of main symbols.}
  \begin{tabular}{cl}
    \toprule
    \textbf{Symbols} & \textbf{Definitions} \\
    \midrule
    $G$ & input attributed network \\
    $\mathbf{A}$ & adjacency matrix \\
    $\mathbf{X}$ & attribute matrix \\
    $\mathcal{T}_t$ &  meta-training task in episode $t$ \\
    $\mathcal{S}_t$ &  support node set in task $\mathcal{T}_t$\\
    $\mathcal{Q}_t$ &  query node set in task $\mathcal{T}_t$\\
    ${\mathbf{W}}$ &  trainable parameter matrix \\
    ${\mathbf h_i^l}$ &   hidden representation of node $v_i$ in  $l^{th}$ layer\\
   	${\mathbf z_i}$ & final latent representation of node $v_i$ \\
   	$\mathbf{p}_c$ & prototype representation of node class $c$ \\
   	$s_i^l$ & importance score of node $v_i$ in  $l^{th}$ layer\\
    $\deg(i)$ & in-degree of node $v_i$ \\
    $C(i)$ & centrality score of node $v_i$\\
    $\Tilde{s}_i$ & centrality-adjusted importance score of node $v_i$\\

    $\hat{y}_i^*$ & predicted class label of query node $v_i^*$ \\
    
    \bottomrule
  \end{tabular}
  \label{table:notation}
\end{table}

\subsection{Few-shot Learning}

Few-shot learning (FSL) aims to solve new tasks with a limited number of examples, based on the knowledge obtained from previous experiences. Generally, existing FSL models fall into two broad categories: (1) \textit{optimization-based approaches}, which focus on learning the optimization of model parameters given the gradients on few-shot examples~\cite{ravi2017optimization,finn2017model,li2017meta,mishra2018simple}. 
One example is the LSTM-based meta-learner~\cite{ravi2017optimization}, which aims to learn efficient parameter updating rules for training a neural classifier. MAML~\cite{finn2017model} learns the parameter initialization that is suitable for different FSL tasks and is compatible with any model trained with gradient descent.
Meta-SGD~\cite{li2017meta} goes further in meta-learning by arguing to learn the weights initialization, gradient update direction and learning rate within
a single step. SNAIL~\cite{mishra2018simple} is another model which combines temporal convolution and soft attention to learn an optimal learning strategy. However, this line of work usually suffers from the computational cost of fine-tuning.
(2) \textit{metric-based approaches}, which try to learn generalizable matching metrics between query and support set across different tasks~\cite{vinyals2016matching,snell2017prototypical,ren2018meta,sung2018learning,liu2019learning}. For instance, Matching Networks~\cite{vinyals2016matching} learn a weighted nearest-neighbor classifier with attention networks. Prototypical Network~\cite{snell2017prototypical} computes the prototype of each class by taking the mean vector of support examples and classifies query instances by calculating their Euclidean distances. An extension of Prototypical Networks proposed by Ren et al.~\cite{ren2018meta} considers both labeled and unlabeled data for few-shot learning. Relation Network~\cite{sung2018learning} trains an auxiliary network to learn a non-linear metric between each query and the support set. It is worth mentioning that our approach also follows this paradigm due to its simplicity and effectiveness. Recently, few-shot learning on graphs has received increasing research attention~\cite{zhou2019meta, bose2019meta}. However, those methods treat support examples equally, rendering the model unstable to noises or outliers~\cite{deng2020meta}.
In this paper, we learn a robust and powerful few-shot learning model by considering the individual importance of labeled support examples. 

%% file: Problem.tex
 \begin{figure*}[t]
    \graphicspath{{figures/}}
    \centering
    \includegraphics[width=0.95\textwidth]{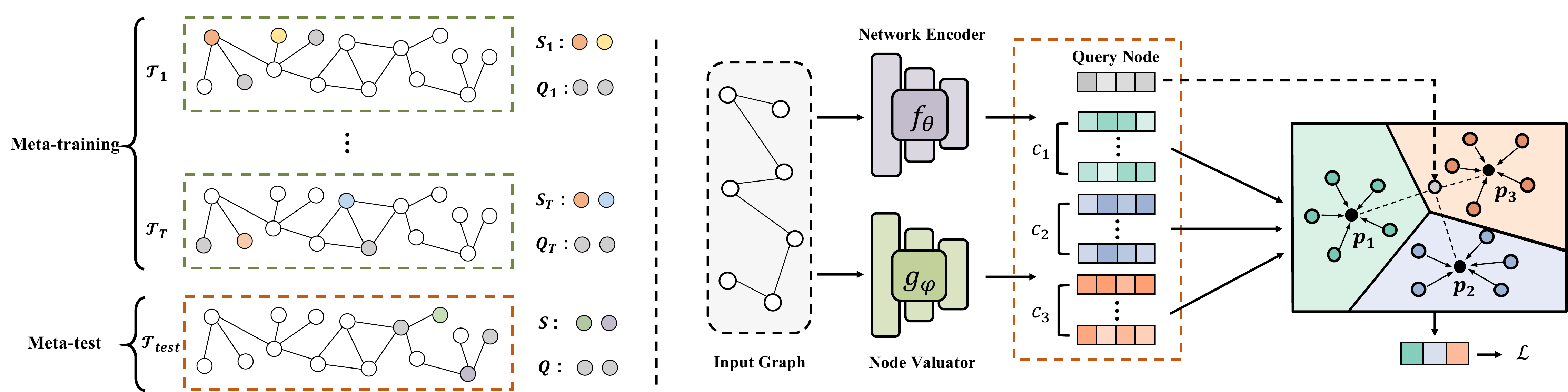}
    \caption{(Left) Episodic training on attributed networks. In each episode, we create a semi-supervised few-shot node classification task by random sampling; (Right) The architecture of the proposed framework Graph Prototypical Networks (GPN).}%
    \label{fig:framework}%
\end{figure*}

Following the commonly used notations, in this paper, we use calligraphic fonts, bold lowercase letters, and bold uppercase letters to denote sets (e.g., $\mathcal{G}$), vectors(e.g., $\mathbf{x}$), and matrices (e.g., $\mathbf{X}$), respectively. The $i^{th}$ row of a matrix $\mathbf{X}$ is denoted by $\mathbf{x}_i$, and the transpose of a matrix $\mathbf{X}$ is represented as $\mathbf{X}^{\mathrm{T}}$. We summarize the main notations used throughout the paper
in Table \ref{table:notation}. For the other special notations, we will illustrate them in the corresponding sections.

Formally, an attributed network can be represented as $G = (\mathcal{V}, \mathcal{E}, \mathbf{X})$, where $\mathcal V$ denotes the set of nodes $\{v_1, v_2, \dots, v_n\}$ and $\mathcal{E}$ denotes the set of edges $\{e_1, e_2, \dots, e_m\}$.
Each node is associated with a
feature vector $\mathbf{x}_i \in \mathbb{R}^{1 \times d}$ and $\mathbf{X} = [\mathbf{x}_1; \mathbf{x}_2; \dots; \mathbf{x}_n] \in \mathbb{R}^{n \times d}$ denotes all the node features. Thus, more generally, the attributed network can be represented as $G = (\mathbf{A}, \mathbf{X})$, where  $\mathbf{A} = \{0, 1\}^{n \times n}$ is an adjacency matrix representing the network structure. Specifically, $\mathbf{A}_{i,j}=1$ indicates that there is an edge between node $v_i$ and node $v_j$; otherwise, $\mathbf{A}_{i,j}=0$. The studied problem can be formulated as follows:

\begin{problem}
\textbf{Few-shot Node Classification on Attributed Networks}:
Given an attributed network $\mathcal{G}=\{\mathbf{A}, \mathbf{X}\}$, suppose we have substantial labeled nodes for a set of node classes $C_{train}$. After training on the labeled data from $C_{train}$, the model is tasked to predict labels for the nodes (i.e., query set $\mathcal{Q}$) from a disjoint set of node classes $C_{test}$, for which only a few labeled nodes of each class (i.e., support set $\mathcal{S}$) are available. 
\end{problem}

Following the common setting in FSL, if $C_{test}$ consists of $N$ classes and the support set $\mathcal{S}$ includes $K$ labeled nodes per class, this problem is named $N$-way $K$-shot node classification problem. In essence, the objective of this problem is to learn a meta-classifier that can be adapted to new classes with only a few labeled nodes. Therefore, how to extract transferable meta-knowledge from $C_{train}$ is the key for solving the studied problem.


%% file: Methodology.tex
As existing FSL models are not tailored for graph-structured data, it is infeasible to apply them to solve the studied problem directly. In this section, we present the details about the proposed Graph Prototypical Networks (GPN) for few-shot node classification on attributed networks. Specifically, our framework is designed and built to address three challenging research questions:
\begin{itemize}[leftmargin=*]
    \item How to perform \textit{meta-learning} on attributed networks (\textit{non-i.i.d.} data) for extracting the meta-knowledge?

    \item How to learn expressive node representations from the input attributed network by considering both the node attributes and topological structure? 
    \item How to identify the informativeness of each labeled node for learning robust and discriminative class representations? 
\end{itemize}

An overview of the proposed Graph Prototypical Networks (GPN) is provided in Figure 2. In Section 4.1, we introduce the backbone training mechanism of the proposed model. In Section 4.2 and 4.3, we introduce how we design the two essential modules in GPN. Then we discuss how to perform few-shot node classification using the proposed framework in Section 4.4. Last, we present the complexity analysis in Section 4.5.



\subsection{Episodic Training on Attributed Networks}
\label{sec:episodic}

Our approach is a meta-learning framework which follows the prevailing episodic training paradigm~\cite{vinyals2016matching}. Specifically, GPN learns over diverse \textit{meta-training} tasks in a large number of episodes rather than only on the target \textit{meta-test} task. The key idea of episodic training is to mimic the real test environment by sampling nodes from $C_{train}$. The consistency between training and test environment alleviates the distribution gap and improves model generalization capability. Specifically, in each episode, we construct a $N$-way $K$-shot meta-training task:
\begin{equation}
    \begin{aligned}
        \mathcal{S}_t &= \{(v_1, y_1),(v_2, y_2), . . . ,(v_{N \times K}, y_{N \times K})\},\\
        \mathcal{Q}_t &= \{(v_1^*, y_1^*),(v_2^*, y_2^*), . . . ,(v_{N \times M}^*, y_{N \times M}^*)\},\\
        \mathcal{T}_t &= \{\mathcal{S}_t, \mathcal{Q}_t\}, 
    \end{aligned}
\end{equation}
where both the support set $\mathcal{S}_t$ and query set $\mathcal{Q}_t$ of the meta-training task $\mathcal{T}_t$ are sampled from $C_{train}$. The support set $ \mathcal{S}_t$ contains $K$ nodes from each class, while the query set $\mathcal{Q}_t$ includes $M$ query nodes sampled from the remainder of each of the $N$ classes. 

The whole training process is based on a set of $T$ meta-training tasks $\mathcal{T}_{train} = \{\mathcal{T}_t\}_{t = 1}^{T}$. The model is trained to minimize the loss of its predictions for the query set $\mathcal{Q}_t$ in each meta-training task $\mathcal{T}_t$, and goes episode by episode until convergence. In this way, the model gradually collects meta-knowledge across those meta-training tasks and then can be naturally generalized to the meta-test task $\mathcal{T}_{test} = \{\mathcal{S}, \mathcal{Q}\}$ with unseen classes $C_{test}$.

Different from conventional episodic training that constructs a pool of supervised \textit{meta-training} tasks~\cite{garcia2017few}, 
in each episode, we sample $N$-way $K$-shot labeled nodes and mask the rest as unlabeled nodes. In this way, we can create a semi-supervised \textit{meta-training} task with the partially labeled attributed network. By considering both labeled and unlabeled data and their dependencies, we are able to learn more expressive node representations for few-shot node classification during the \textit{meta-learning} process.





\subsection{Network Representation Learning}

In order to learn expressive node representations from an attributed network, we develop a \textit{network encoder} to capture the data heterogeneity. Specifically, the \textit{network encoder} possesses a GNN backbone, which converts each node to a low-dimensional latent representation. In general, GNNs follow the neighborhood aggregation scheme, and compute the node representations by recursively aggregating and compressing node features from local neighborhoods. Briefly, a GNN layer can be defined as:

\begin{equation}
\begin{aligned}
    \mathbf{h}_i^l &= \textsc{Combine}^l\Big( \mathbf{h}_i^{l-1},  \mathbf{h}_{\mathcal{N}_i}^l\Big),\\
    \mathbf{h}_{\mathcal{N}_i}^l &= \textsc{Aggregate}^l\Big(\{   \mathbf{h}_j^{l-1} \arrowvert \forall j \in \mathcal{N}_i \cup v_i \}\Big),
    \label{eq:graphSage}
\end{aligned}
\end{equation}
where $\mathbf{h}_i^{l}$ is the node representation of node $i$ at layer $l$ and $\mathcal{N}_i$ is the set of neighboring nodes of $v_i$. \textsc{Combine} and \textsc{Aggregate} are two key functions of GNNs and have a series of possible implementations~\cite{kipf2016semi,hamilton2017inductive,velickovic2017graph}.



 By stacking multiple GNN layers in the \textit{network encoder}, the learned node representations are able to capture the long-range node dependencies in the network:
\begin{equation}
\begin{aligned}
    &\mathbf{H}^{1} = \text{GNN}^1 (\mathbf{A}, \mathbf{X}),\\
    &\dots\\
    &\mathbf{Z} = \text{GNN}^L (\mathbf{A}, \mathbf{H}^{L-1}),
\end{aligned}
\end{equation}
where $\mathbf{Z}$ is the learned node representations from the \textit{network encoder}. For simplicity, we will use $f_{\bm{\theta}}(\cdot)$ to denote the \textit{network encoder}
with $L$ GNN layers.

\smallskip
\noindent\textbf{Prototype Computation.}
With the learned node representations from the \textit{network encoder}, next, we aim to compute the representation of each class with the labeled nodes from the support set. We follow the idea of Prototypical Networks~\cite{snell2017prototypical}, which encourages nodes of each class cluster around a specific prototype representation. Formally, the class prototypes can be computed by:
\begin{equation}
    \mathbf{p}_c = \textsc{Proto} \Big( \{  \mathbf{z}_i | \forall i \in \mathcal{S}_c \}\Big),
\end{equation}
where $\mathcal{S}_c$ denotes the set of labeled examples from class $c$ and $\textsc{Proto}$ is the prototype computation function. For instance, in the vanilla Prototypical Networks~\cite{snell2017prototypical}, the prototype of each class is computed by taking the average of all embedded nodes belonging to that class:
\begin{equation}
    \mathbf{p}_c = \frac{1}{|\mathcal{S}_c|} \sum_{i \in \mathcal{S}_c} \mathbf{z}_i.
\end{equation}

\subsection{Node Importance Valuation}
Despite its simpleness, directly taking the mean vectors of the embedded support instances as prototypes may not provide promising results for our problem. It not only neglects the fact that each node has a different significance in a network, but also makes the FSL model highly noise-sensitive since labeled data is severely limited~\cite{zhang2019variational}. Therefore, refining those class prototypes becomes especially essential for building a robust and effective FSL model. 

To identify the informativeness of each labeled node, we adopt a view that the importance of a node is highly correlated with its neighbors' importance~\cite{park2019estimating}. Accordingly, we design a GNN-based \textit{node valuator} $g_\phi(\cdot)$ (as shown in Figure 3) to estimate node importance scores through a \textit{score aggregation layer}, which can be defined as follows:



\begin{equation}
    s^l_i = \sum_{j \in \mathcal{N}_i \cup v_i} \alpha_{ij}^l s^{l-1}_j,
\end{equation}
where $s_i^l$ is the importance score of node $v_i$ in the $l$-th layer ($l = 1, \dots , L$). $\alpha_{ij}^l$ is the attention weight between nodes $v_i$ and $v_j$, we compute it via a shared attention mechanism: 
\begin{equation}
    \alpha_{ij}^l = \frac{\text{exp}\big(\text{LeakyReLU}\big(\mathbf{a}^{\mathrm{T}}[s_i^{l-1} || s_j^{l-1}]\big)\big)}{\sum_{k \in \mathcal{N}_i \cup v_i} \text{exp}\big(\text{LeakyReLU}\big(\mathbf{a}^{\mathrm{T}} [s_j^{l-1} || s_k^{l-1}]\big)\big)},
\end{equation}
where $||$ is a concatenation operator and $\mathbf{a}$ is a weight vector.

To compute the initial importance score $s^0_i$, we employ a
\textit{scoring layer} to compress the node features. Our \textit{scoring layer} is a feed-forward layer with tanh non-linearity. Specifically, the initial score of node $v_i$ is computed by:
\begin{equation}
    s_i^0 = \text{tanh}(\mathbf{w}^{\mathrm{T}}_s \mathbf{x}_i + b_s)
\end{equation}
where $\mathbf{w}_s \in \mathbb{R}^d$ is a learnable weight vector and $b_s \in \mathbb{R}^1$ is the bias. 


\smallskip
\noindent\textbf{Centrality Adjustment.} As suggested in previous research on node importance estimation~\cite{page1999pagerank,park2019estimating} , the importance of a node positively correlates with
its centrality in the graph. Given that the in-degree $\deg(i)$ of node $v_i$ is a common proxy for its centrality and popularity, we define the initial centrality $C(i)$ of node $v_i$ as:
\begin{equation}
    C(i) = \log(\deg(i) + \epsilon),
\end{equation}
where $\epsilon$ is a small constant. To compute the final importance score, we apply centrality adjustment to the estimated score $s_i^L$ from the last layer, and apply a sigmoid non-linearity as follows:
\begin{equation}
    \Tilde{s}_i = \text{sigmoid}(C(i) \cdot s^{L}_i).
\end{equation}

\begin{figure}[t]
    \graphicspath{{figures/}}
    \centering
    \includegraphics[width=0.825\linewidth]{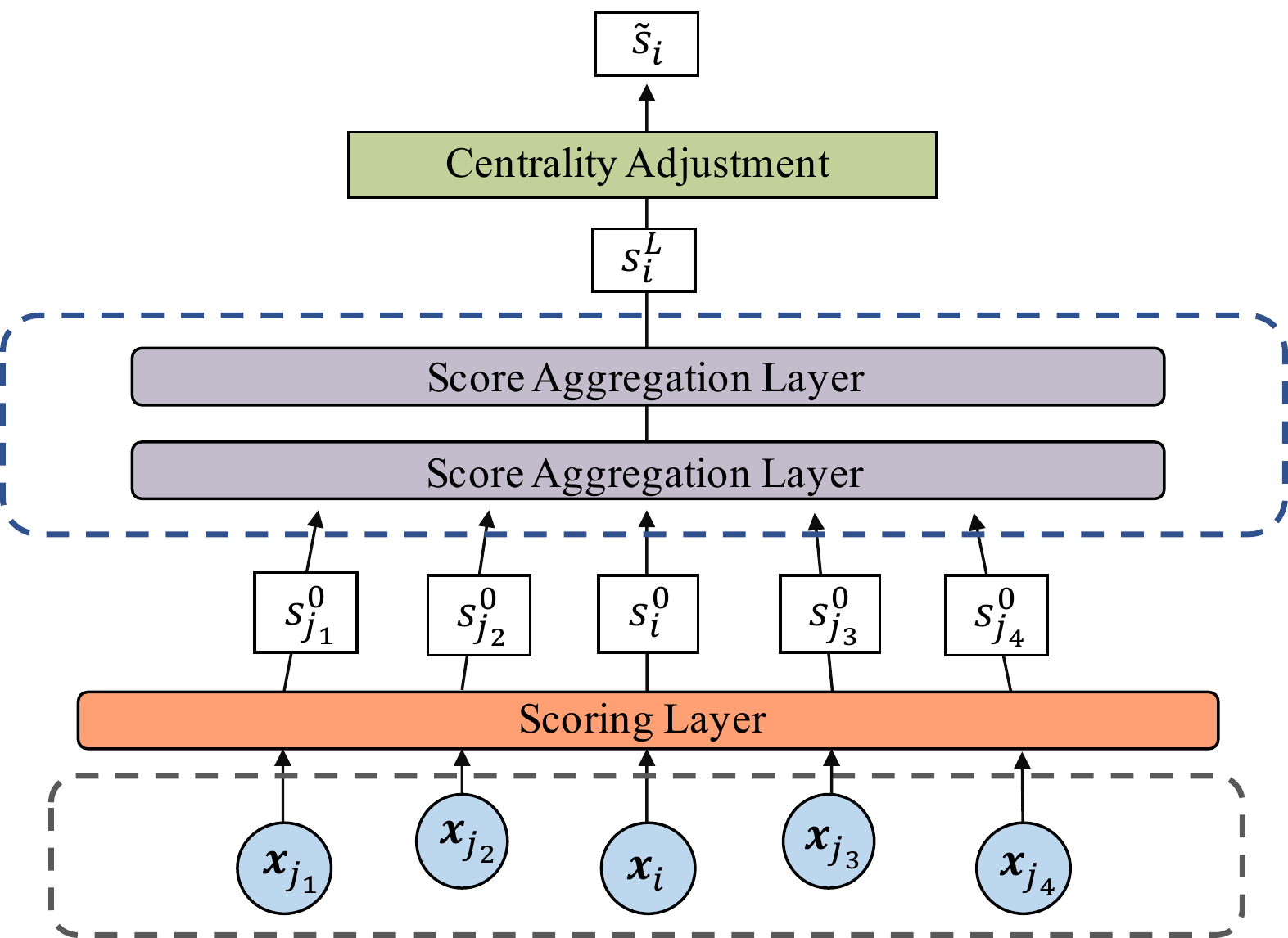}
    \caption{Architecture of the node valuator.}%
    \label{fig:estimator}%
\end{figure}

In this way, the \textit{node valuator} adjusts the importance of labeled examples in the support set by making use of the additional information encoded in the network. 



\begin{algorithm}[t]
\caption{Learning process of GPN.}
\LinesNumbered
\small
\KwIn{Attributed network $G = (\mathbf{A}, \mathbf{X})$, few-shot node classfication task $\mathcal{T}_{test} = \{\mathcal{S}$, $\mathcal{Q}\}$, training episodes $T$. }
\KwOut{Predicted labels of nodes in the query set $\mathcal{Q}$.}
// \texttt{Meta-training process}

\While{$i < T$}{

    Sample a meta-training task $\mathcal{T}_i = \{\mathcal{S}_i, \mathcal{Q}_i\}$

    
     
     Compute representations for the nodes in $\mathcal{S}_i$ and $\mathcal{Q}_i$\;
     Estimate importance scores for the nodes in $\mathcal{S}_i$\;
     
     Minimize the meta-training loss according to Eq. (\ref{eq:loss})\;
    }
// \texttt{Meta-test process}   

    Compute representations for the nodes in $\mathcal{S}$ and $\mathcal{Q}$\;
    Estimate importance scores for the nodes in $\mathcal{S}_i$\;
    
    Predict labels for the nodes in the query set $\mathcal{Q}$\;
\end{algorithm}

\subsection{Few-shot Node Classification}

After we compute the importance score of each support node, we first normalize those scores using the softmax function:
\begin{equation}
    \beta_i = \frac{\text{exp}(\Tilde{s}_i)}{\sum_{k \in \mathcal{S}_c}  \text{exp}(\Tilde{s}_k)},
\end{equation} 
where $\beta_i$ represents the normalized weight of each support node $v_i$, then the refined prototypes can be directly computed by:
\begin{equation}
        \mathbf{p}_c =  \sum_{i \in \mathcal{S}_c}\beta_i  \mathbf{z}_i.
\label{eq:prototype}
\end{equation}

As such, our model can adjust the cluster locations to better represent the examples in both the support and unlabeled sets. These learned prototypes define a predictor for the class label of a query node $v_i^*$, which assigns a probability over each class $c$ based on the distances between the query node $v_i^*$ and each prototype:
\begin{equation}
    p(c | v_i^*) = \frac{\exp(-d( \mathbf{z}_i^*, \mathbf{p}_c))}{\sum_{c'}\exp(-d( \mathbf{z}_i^*, \mathbf{p}_{c'}))},
\end{equation}
where $d(\cdot)$ is a distance metric function. Commonly, squared Euclidean distance is a simple and effective choice~\cite{snell2017prototypical}.

Under the episodic training framework, the objective of each meta-training task is to minimize the classification loss between the predictions of the query set and the ground-truth. Specifically, the training loss can be defined as the average negative log-likelihood probability of assigning correct class labels:
\begin{equation}
    \mathcal{L} = - \frac{1}{N \times M} \sum_{i=1}^{N \times M} \log p(y_i^* | v_i^*).
\label{eq:loss}
\end{equation}

By minimizing the above loss function, GPN is able to learn a generic classifier for a specific meta-training task. Training episodes are formed by randomly selecting a subset of classes from the auxiliary class set $C_{train}$, then choosing a subset of nodes within each class to act as the support set and a subset of the remainder to serve as query set. After training on a considerable number of meta-training
tasks, its generalization performance will be measured on the test episodes, which contain nodes sampled from $\mathcal{C}_{test}$ instead of $\mathcal{C}_{train}$. For each test episode, we use the predictor produced by our GPN for the provided support set $\mathcal{S}$ to classify each query node in $\mathcal{Q}$ into the most likely class:
$\hat{y}_i^* = \text{argmax}_c p(c|v^*_i
)$. The detailed learning process of GPN is presented in Algorithm 1.

\subsection{Complexity Analysis}

Our proposed framework GPN is composed of two main components introduced in the previous sections. As both the \textit{network encoder} and \textit{node valuator} are built upon graph neural networks, the complexity of GPN mainly depends on the specific underlying GNN architecture. For instance, the computational complexity of a GCN layer is $O(|\mathcal{E}| d d')$, where $|\mathcal{E}|$ denotes the number of edges in the attributed network, $d$ and $d'$ are the input feature size and output feature size, respectively~\cite{kipf2016semi}. Note the complexity of the scoring layer is $O(|\mathcal{V}|dd')$ and score aggregation layer is $O(|\mathcal{V}| + |\mathcal{E}|)$, where $|\mathcal{V}|$ denotes the number of nodes in the network. Overall, as $|\mathcal{E}| \gg |\mathcal{V}|$ in practice, the complexities of GPN models can be considered as linear with respect to the number of edges.

%% file: Experiments.tex
In order to verify the effectiveness of our proposed framework, in this section, we first introduce the experimental settings and then present the detailed experiment results\footnote{Code and data are available at \url{https://github.com/kaize0409/GPN}}.

\begin{table}[t]
\caption{Statistics of the evaluation datasets.}
\centering
\setlength{\tabcolsep}{3pt}
\begin{tabular}{@{}lcccc@{}}
\toprule

\textbf{Datasets} & \# nodes & \# edges  & \# attributes & \# labels  \\ \midrule
Amazon-Clothing & 24,919  & 91,680   & 9,034 & 77\\
Amazon-Electronics & 42,318  & 43,556   & 8,669 & 167\\
DBLP & 40,672 & 288,270   & 7,202   & 137 \\
Reddit & 232,965  & 11,606,919    & 602 & 41  \\
\bottomrule
\end{tabular}
\label{tab:datasets}
\end{table}

\begin{table*}[t!]
\centering

\caption{Averaged few-shot node classification results on four datasets w.r.t ACC and F1 (\%).}

\setlength{\tabcolsep}{2.5pt}
\scalebox{0.965}{
\begin{tabular}{@{}lccccccccccc cccccccccccc@{}}
\toprule

\rule{0pt}{10pt} & \multicolumn{11}{c}{\textbf{Amazon-Clothing}} & & \multicolumn{11}{c}{\textbf{Amazon-Electronics}}\\ \cline{2-12} \cline{14-24}

\rule{0pt}{10pt}  \multirow{2}{*}{
\textbf{Methods}} & \multicolumn{2}{c}{ 5-way 3-shot}  & &  \multicolumn{2}{c}{ 5-way 5-shot}  &  & \multicolumn{2}{c}{ 10-way 3-shot}  &  & \multicolumn{2}{c}{ 10-way 5-shot}  &&
\multicolumn{2}{c}{ 5-way 3-shot}  & &  \multicolumn{2}{c}{ 5-way 5-shot}  &  & \multicolumn{2}{c}{ 10-way 3-shot}  &  & \multicolumn{2}{c}{ 10-way 5-shot}

\\ \cline{2-3} \cline{5-6} \cline{8-9} \cline{11-12}
\cline{14-15} \cline{17-18} \cline{20-21} \cline{23-24}

\rule{0pt}{10pt} & \multicolumn{1}{c}{ACC}  & \multicolumn{1}{c}{F1} & & \multicolumn{1}{c}{ACC} & \multicolumn{1}{c}{F1} & &
\multicolumn{1}{c}{ACC} & \multicolumn{1}{c}{F1}  & &
\multicolumn{1}{c}{ACC} & \multicolumn{1}{c}{F1} &&

\multicolumn{1}{c}{ACC}  & \multicolumn{1}{c}{F1} & & \multicolumn{1}{c}{ACC} & \multicolumn{1}{c}{F1} & &
\multicolumn{1}{c}{ACC} & \multicolumn{1}{c}{F1}  & &
\multicolumn{1}{c}{ACC} & \multicolumn{1}{c}{F1}

\\ \hline

DeepWalk        & 36.7  & 36.3 & & 46.5 & 46.6 & & 21.3 & 19.1 & & 35.3 & 32.9 && 23.5 & 22.2 && 26.1 & 25.7 && 14.7 & 12.9 && 16.0 & 14.7    \\
node2vec        & 36.2 & 35.8 & & 41.9 & 40.7 & & 17.5 & 15.1 && 32.6 & 30.2 && 25.5 & 23.7 && 27.1 & 24.3 && 15.1 & 13.1 && 17.7 & 15.5\\

GCN      &  54.3 & 51.4 && 59.3 & 56.6 && 41.3 & 37.5 && 44.8 & 40.3 && 53.8 & 49.8 && 59.6 & 55.3 && 42.3 & 38.4 && 47.4 & 48.3 \\

SGC & 56.8 & 55.2 && 62.2 & 61.5 && 43.1 & 41.6 && 46.3 & 44.7 && 54.6 & 53.4 && 60.8 & 59.4 && 43.2 & 41.5 && 50.0 & 47.6

\\
\midrule

PN  & 53.7 & 53.6 && 63.5 & 63.7 && 41.5 & 41.9 && 44.8 & 46.2 && 53.5 & 55.6 && 59.7 & 61.5 && 39.9 & 40.0 && 45.0 & 44.8   \\
MAML  & 55.2 & 54.5 && 66.1 & 67.8 && 45.6 & 43.3 && 46.8 & 45.6 && 53.3 & 52.1 && 59.0 & 58.3 && 37.4 & 36.1 && 43.4 & 41.3   \\

Meta-GNN  & 74.1 & 73.6 && 77.3 & 77.5 && 61.4 & 59.7 && 64.2 & 62.9  && 63.2 & 61.5 && 67.9 & 66.8 && 58.2 & 55.8 && 60.8 & 60.1     \\

GPN   & \textbf{75.4} & \textbf{74.7} &&  \textbf{78.6} & \textbf{79.0} && \textbf{65.0} & \textbf{66.1} && \textbf{67.7} & \textbf{68.9} && \textbf{64.6} & \textbf{62.8} && \textbf{70.9} & \textbf{70.6} && \textbf{60.3} & \textbf{60.7} && \textbf{62.4} & \textbf{63.7}\\

\bottomrule
\end{tabular}}

\bigskip
\setlength{\tabcolsep}{2.5pt}
\scalebox{0.965}{
\begin{tabular}{@{}lccccccccccc cccccccccccc@{}}
\toprule

\rule{0pt}{10pt} & \multicolumn{11}{c}{\textbf{DBLP}} & & \multicolumn{11}{c}{\textbf{Reddit}}\\ \cline{2-12} \cline{14-24}

\rule{0pt}{10pt}  \multirow{2}{*}{\textbf{Methods}} & \multicolumn{2}{c}{ 5-way 3-shot}  & &  \multicolumn{2}{c}{ 5-way 5-shot}  &  & \multicolumn{2}{c}{ 10-way 3-shot}  &  & \multicolumn{2}{c}{ 10-way 5-shot}  &&
\multicolumn{2}{c}{ 5-way 3-shot}  & &  \multicolumn{2}{c}{ 5-way 5-shot}  &  & \multicolumn{2}{c}{ 10-way 3-shot}  &  & \multicolumn{2}{c}{ 10-way 5-shot}

\\ \cline{2-3} \cline{5-6} \cline{8-9} \cline{11-12}
\cline{14-15} \cline{17-18} \cline{20-21} \cline{23-24}

\rule{0pt}{10pt} & \multicolumn{1}{c}{ACC}  & \multicolumn{1}{c}{F1} & & \multicolumn{1}{c}{ACC} & \multicolumn{1}{c}{F1} & &
\multicolumn{1}{c}{ACC} & \multicolumn{1}{c}{F1}  & &
\multicolumn{1}{c}{ACC} & \multicolumn{1}{c}{F1} &&

\multicolumn{1}{c}{ACC}  & \multicolumn{1}{c}{F1} & & \multicolumn{1}{c}{ACC} & \multicolumn{1}{c}{F1} & &
\multicolumn{1}{c}{ACC} & \multicolumn{1}{c}{F1}  & &
\multicolumn{1}{c}{ACC} & \multicolumn{1}{c}{F1}

\\ \hline

DeepWalk        & 44.7  & 43.1 & & 62.4 & 60.4 & & 33.8 & 30.8 & & 45.1 & 43.0  && 26.7 & 26.1 && 30.1 & 29.7 && 17.6 & 17.1 && 18.8 & 18.6   \\
node2vec        & 40.7 & 38.5 & & 58.6 & 57.2 & & 31.5 & 27.8 && 41.2 & 39.6 && 27.1 & 25.6 && 31.2 & 29.8 && 19.8 & 18.6 && 23.4 & 22.6 \\

GCN        &  59.6 & 54.9 && 68.3 & 66.0 && 43.9 & 39.0 && 51.2 & 47.6  && 38.8 & 38.1 && 45.5 & 44.1 && 29.0 & 27.0 && 35.7 & 32.4  \\
SGC & 57.3 & 54.7 && 65.0 & 62.1 && 40.2 & 36.8 && 50.3 & 46.4 && 44.4 & 42.1 && 46.8 & 42.5 && 29.7 & 26.8 && 31.6 & 27.7  \\
\midrule
PN  & 37.2 & 36.7 && 43.4 & 44.3 && 26.2 & 26.0 && 32.6 & 32.8 && 34.6 & 33.3 && 37.6 & 36.4 && 19.8 & 18.0 && 23.3 & 21.4  \\
MAML  & 39.7 & 39.7 && 45.5 & 43.7 && 30.8 & 25.3 && 34.7 & 31.2 && 29.1 & 26.8 && 31.1 & 29.7 && 15.2 & 12.2 && 17.9 & 15.6  \\
Meta-GNN  & 70.9 & 70.3 && 78.2 & 78.2 && 60.7 & 60.4 && 68.1 & 67.2  && 60.8 & 58.3 && 62.7 & 61.2 && 44.9  & 42.1 && 51.5  & 47.1     \\
GPN   &  \textbf{74.5} & \textbf{73.9} &&  \textbf{80.1} & \textbf{79.8} && \textbf{62.6} & \textbf{62.6} && \textbf{69.0} & \textbf{69.4} && \textbf{65.5} & \textbf{66.2} && \textbf{68.4} & \textbf{69.0} && \textbf{53.4} & \textbf{55.8} && \textbf{57.7} & \textbf{59.2}\\

\bottomrule
\end{tabular}}
\label{table:semi}
\vspace{0.15cm}
\end{table*}

\subsection{Experiment Settings}
\label{sec:setting}
\smallskip
 \noindent{\textbf{Evaluation Datasets.}} Due to the fact that few-shot node classification on graph-structured data remains an under-studied problem, it is worth mentioning that the existing benchmark datasets (e.g., Cora, Pubmed) for conventional node classification problem are not suitable for evaluating FSL models. The main reason is that FSL models usually need to be tested on many different classification tasks, while those datasets only contain limited node classes. To extensively evaluate the model performance on few-shot node classification, in our experiments, we adopt four public datasets with plenty of node classes, including:

\begin{itemize}[leftmargin=*]
    \item \textbf{Amazon-Clothing}~\cite{mcauley2015inferring} is a product network built with the products in ``Clothing, Shoes and Jewelry'' on Amazon. In this dataset, each product is considered as a node and its description is used to construct the node attributes. We use the substitutable relationship (``also viewed'') to create links between products. The class label is defined as the low-level product category. For this dataset, we use 40/17/20 node classes for training/validation/test. 
    
    \item \textbf{Amazon-Electronics}~\cite{mcauley2015inferring} is another Amazon product network which contains products belonging to ``Electronics''. Similar to the first dataset, each node denotes a product and its attributes represent the product description. Note that here we use the complementary relationship (``bought together'') between products to create the edges. The low-level product categories are used as class labels. For this dataset, we use 90/37/40 node classes for training/validation/test.
        
    \item \textbf{DBLP}~\cite{tang2008arnetminer} is a citation network where each node represents a paper, and the links are the citation relations among different papers. The paper abstracts are used to construct node attributes. The class label of a node is defined as the paper venue. For this dataset, we use 80/27/30 node classes for training/validation/test.
    
    \item \textbf{Reddit}~\cite{hamilton2017inductive} is a post-to-post graph constructed with data sampled from Reddit, which is used to evaluate the performance of our model on large-scale attributed networks. In this large-scale attributed network, posts are represented by nodes and two posts are connected if they are commented by the same user. Each post is labeled with it a community ID. For this dataset, we use 16/10/15 node classes for training/validation/test.

\end{itemize}

We summarize the statistics of the above datasets in Table \ref{tab:datasets}. More details, such as data sources and how they are preprocessed, can be found in Appendix A.1.

\smallskip
\noindent{\textbf{Compared Methods.}}
In the experiments, we compare the proposed model GPN with related baseline methods, including:
\begin{itemize}[leftmargin=*]
     \item \textbf{DeepWalk}~\cite{perozzi2014deepwalk}: It performs a stream of truncated vanilla random walks on the input graph, and learns node embeddings from the sampled random walks.
        
    \item \textbf{node2vec}~\cite{grover2016node2vec}: It extends DeepWalk with biased random walks to explore diverse neighborhoods.
    
    \item \textbf{GCN}~\cite{kipf2016semi}: This model learns latent node representations based on the first-order approximation of spectral graph convolutions.
    \item \textbf{SGC}~\cite{wu2019simplifying}: It reduces the extra complexity of GCN by eliminating the non-linearity between the GCN layers and folding the convolution functions into a linear transformation.
    \item \textbf{PN}~\cite{snell2017prototypical}: Prototypical Network is one of the widely used few-shot learning methods for image classification.
    \item \textbf{MAML}~\cite{finn2017model}: It is an optimization-based meta-learning method, which tries to learn a better model initialization from a series of meta-training tasks.
    \item \textbf{Meta-GNN}~\cite{zhou2019meta}: This baseline extends MAML to graph data by using a GNN base model.
\end{itemize}

The above baseline methods can be summarized into three categories: (1) \textit{random walk-based methods} including two widely used unsupervised methods DeepWalk and node2vec. With the learned node representations, we train a Logistic Regression~\cite{kleinbaum2002logistic} classifier to perform node classification; (2) \textit{GNN-based methods} including two state-of-the-art models GCN and SGC for semi-supervised node classification. (3) \textit{few-shot methods} including PN, MAML and Meta-GNN. Specifically, PN and MAML are two representative few-shot learning models for \textit{i.i.d.} data, while Meta-GNN is able to handle graph-structured data by integrating graph neural networks with meta-learning. Note that for the first two categories of methods, we follow the way in ~\cite{zhou2019meta} to adapt those models to few-shot node classification scenarios. 


\smallskip
\noindent\textbf{Implementation of GPN.} We implement the proposed framework in PyTorch and the code is public. Specifically, the \textit{network encoder} consists of two GCN layers with dimension size 32 and 16, respectively. Both of them are activated with ReLU function. For the \textit{node valuator}, it consists of one fully-connected layer and two score aggregation layers. For each score aggregation layer, we use Leaky ReLU with a negative
slope of 0.2 as the activation function. GPN is trained with Adam optimizer, whose learning rate is set to be $\alpha = 0.005$ initially with a weight decay of $0.0005$. The coefficients for computing running averages of gradient and square are set to be $\beta_1 = 0.9, \beta_2 = 0.999$. To avoid overfitting, we finetune the dropout rate and determine the value for each dataset based on the validation performance. For each dataset, we train the model over 300 episodes with an early-stopping strategy.

\subsection{General Comparisons}
For each dataset, we evaluate the performance of all the algorithms on four few-shot node classification tasks, i.e., $5$-way-$3$-shot, $5$-way-$5$-shot, $10$-way-$3$-shot, and $10$-way-$5$-shot. We set the query size as same as the support size in our experiments. We adopt two widely used metrics Accuracy (ACC) and Micro-F1 (F1) to evaluate performance. Each model is evaluated on 50 \textit{meta-test} tasks and each \textit{meta-test} task is randomly sampled from test node classes. We repeat the process 10 times and the averaged results are presented in Table \ref{table:semi}. Higher values are better for all metrics. From the comprehensive views, we make the following observations: 

\begin{itemize}[leftmargin=*]
    \item A general observation is that our approach GPN achieves the best performance on all the few-shot tasks. For example, on the Amazon-Clothing dataset, GPN outperforms the best performing baseline Meta-GNN by 5.9\% (ACC) under the $10$-way-$3$-shot task. The improvements are even more substantial on the larger dataset Reddit. This result verifies that GPN is a powerful and reliable model to tackle the problem of few-shot node classification on attributed networks.


    \item Overall, DeepWalk and node2vec largely fall behind other methods on few-shot node classification tasks. Those random walk-based methods need to train a supervised classifier (e.g., Logistic Regression) with learned node representations, which typically rely on a large number of labeled data for good performance. Similarly, GNN-based methods are unable to obtain competitive results on the few-shot node classification problem. Conventional GNN models are developed for semi-supervised node classification, and could be easily overfitted with only a small number of labeled instances.
    
    \item Despite the success of MAML and PN on few-shot image classification, however, both of them perform poorly on our tasks.
    The main reason is that those methods cannot capture the dependency between nodes for learning expressive node representations, rendering unsatisfactory performance on few-shot node classification tasks.

    \item By integrating the idea of meta-learning into graph neural networks, Meta-GNN is able to achieve considerable improvements over other baseline methods on few-shot node classification in most cases. However, it is worth noting that its performance suffers a catastrophic decline on the Reddit dataset. One reasonable explanation is that optimization-based FSL approaches require extensive fine-tuning efforts for the target task, especially on those large-scale datasets.

\end{itemize}

\begin{figure*}[!t]
    
    \graphicspath{{figures/}}
    \centering
    \scalebox{0.975}{
    \subfigure[\textbf{Amazon-Clothing}]
    {
    \hspace{-0.2cm}
    \includegraphics[width=0.23\textwidth]{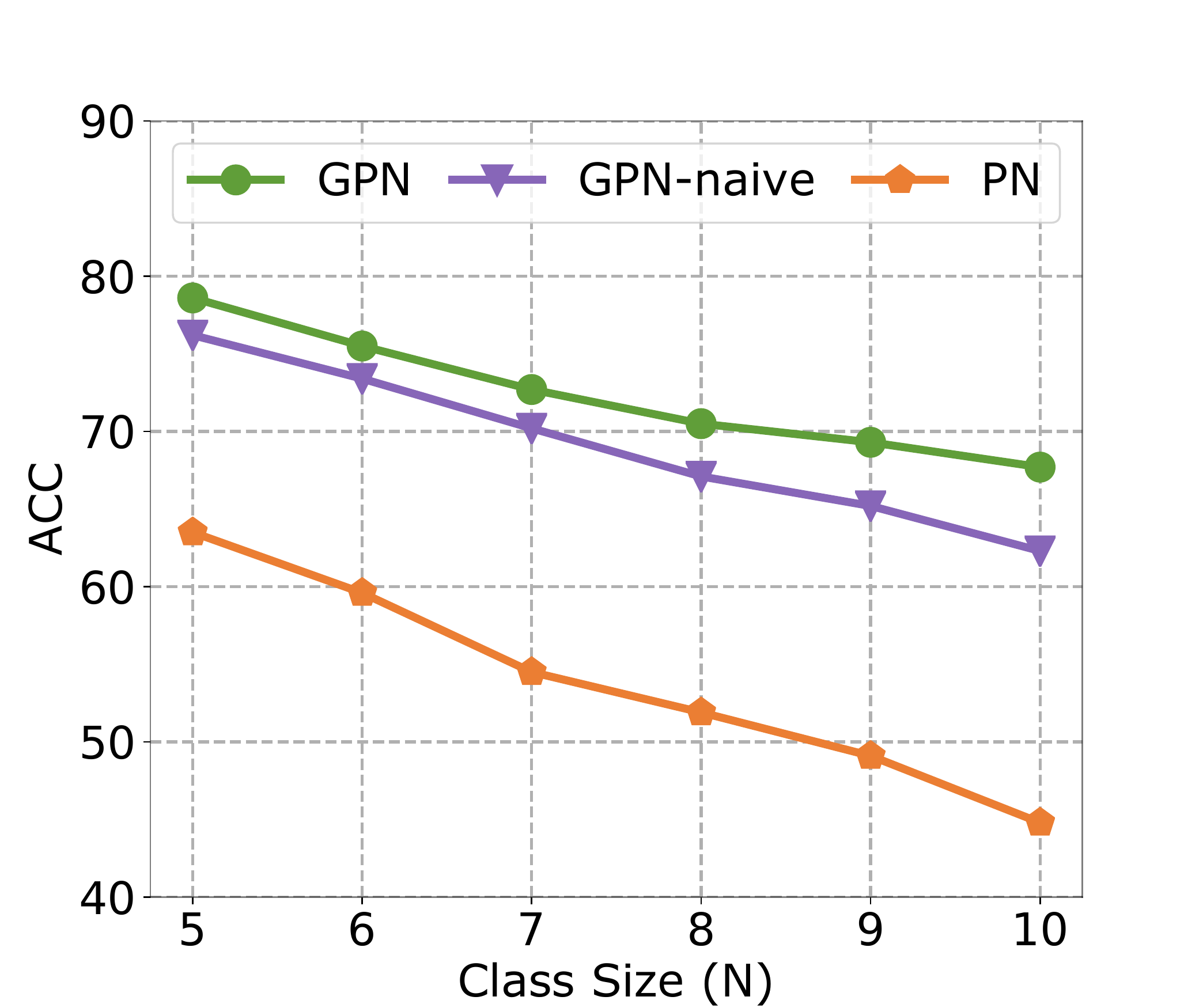}
    }
    \hspace{-0.2cm}
    \subfigure[\textbf{Amazon-Electronics}]
    {
    \includegraphics[width=0.23\textwidth]{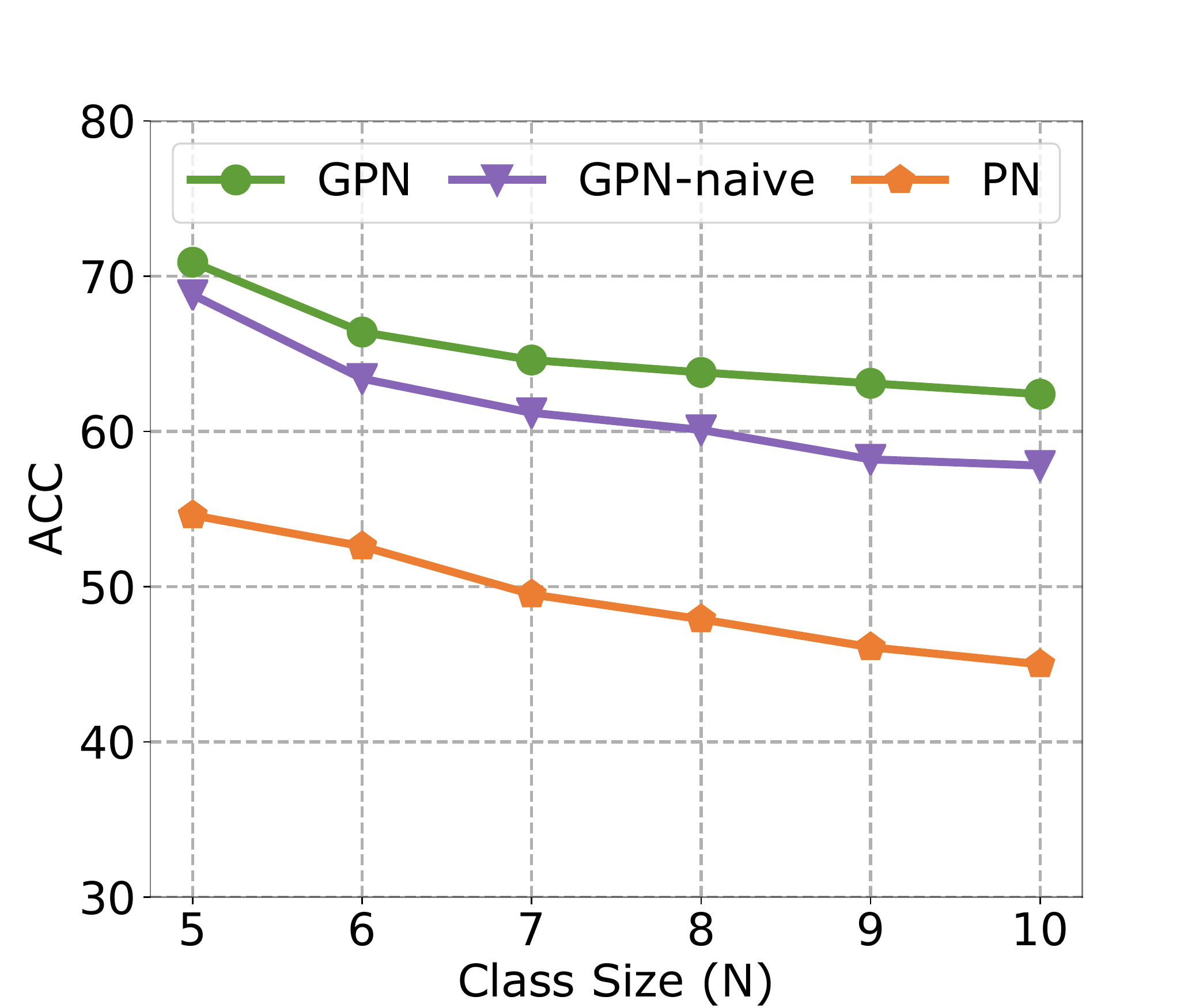}
    }
    \hspace{-0.2cm}
    \subfigure[\textbf{DBLP}] 
    {
    \includegraphics[width=0.23\textwidth]{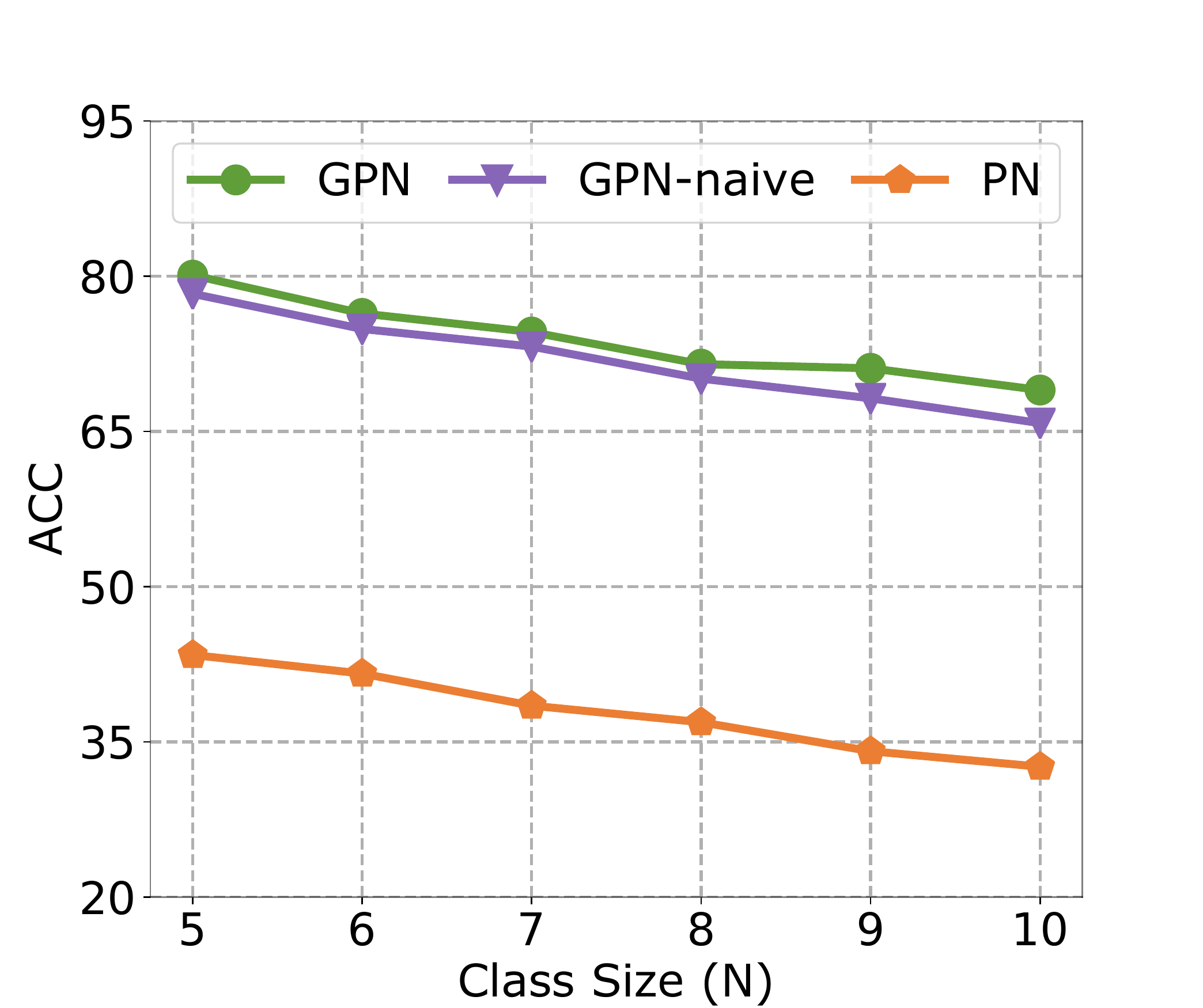}
    }
    \hspace{-0.2cm}
    \subfigure[\textbf{Reddit}]
    {
    \includegraphics[width=0.23\textwidth]{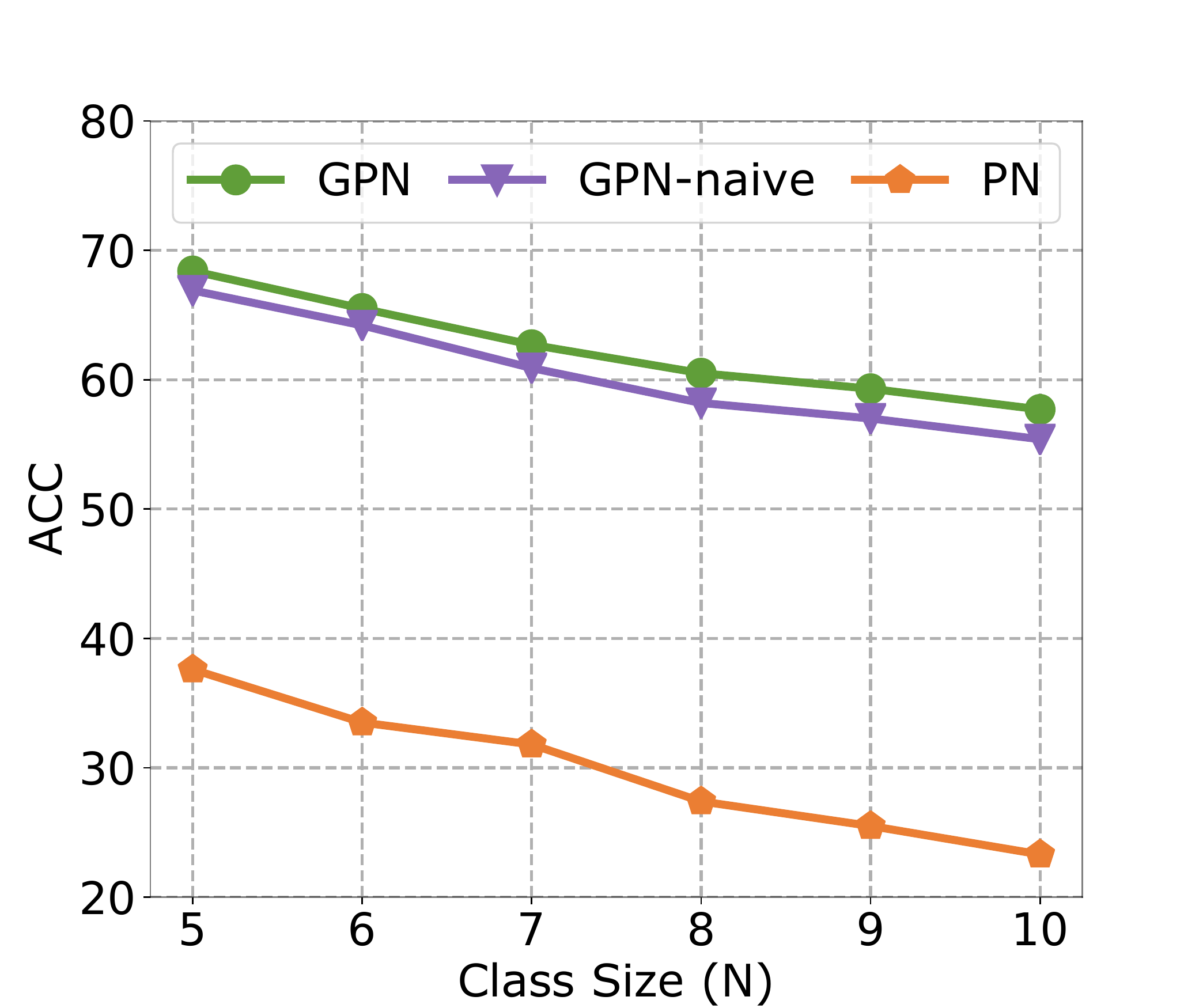}
    }}
    \caption{Performance comparisons \textit{w.r.t.} test class size ($N$-way $5$-shot).}%
    \label{fig:N}
\end{figure*}  

\begin{figure*}[!t]
    \graphicspath{{figures/}}
    \centering
    \scalebox{0.975}{
    \subfigure[\textbf{Amazon-Clothing}]
    {
    \hspace{-0.2cm}
    \includegraphics[width=0.23\textwidth]{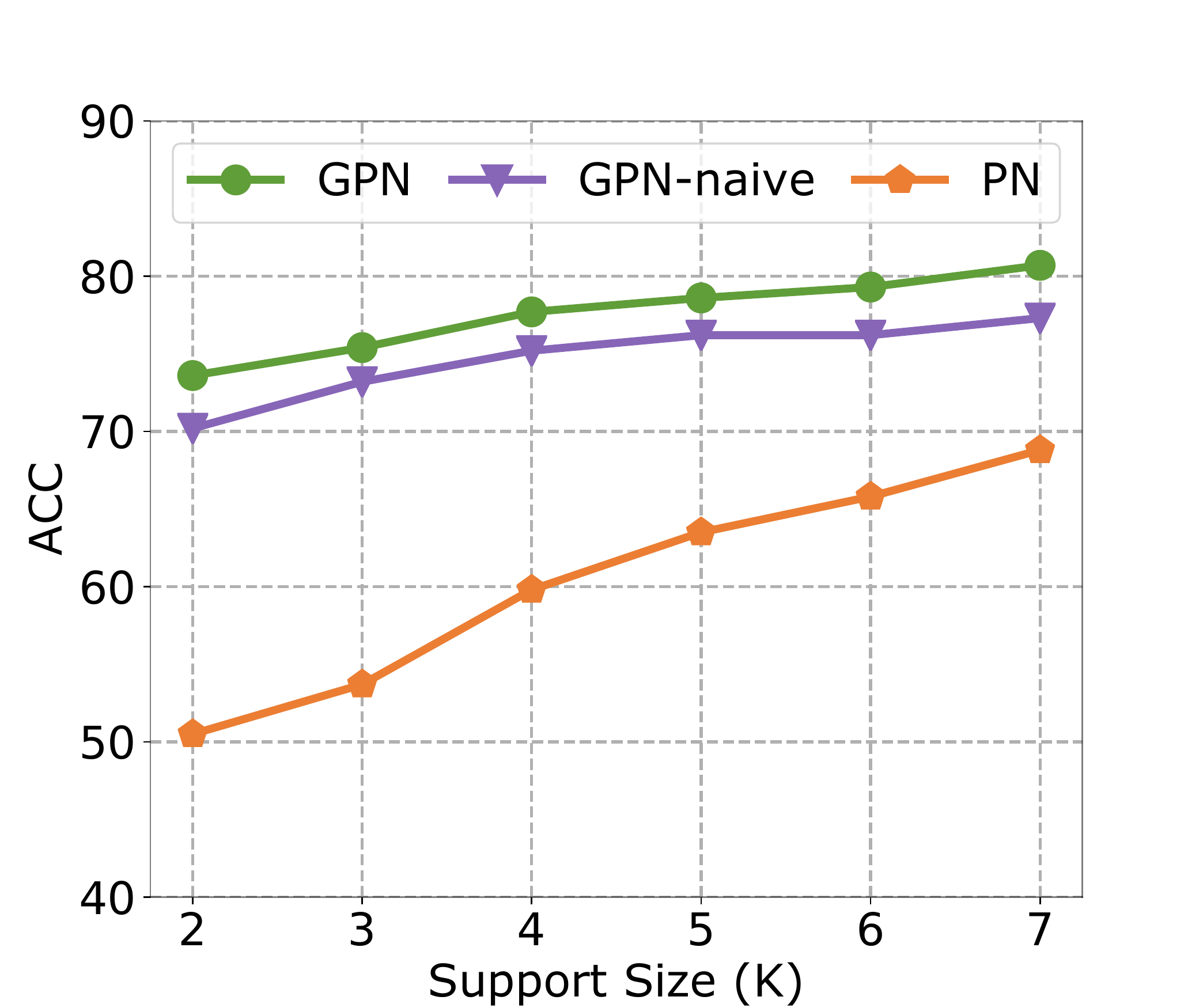}
    }
    \hspace{-0.2cm}
    \subfigure[\textbf{Amazon-Electronics}]
    {
    \includegraphics[width=0.23\textwidth]{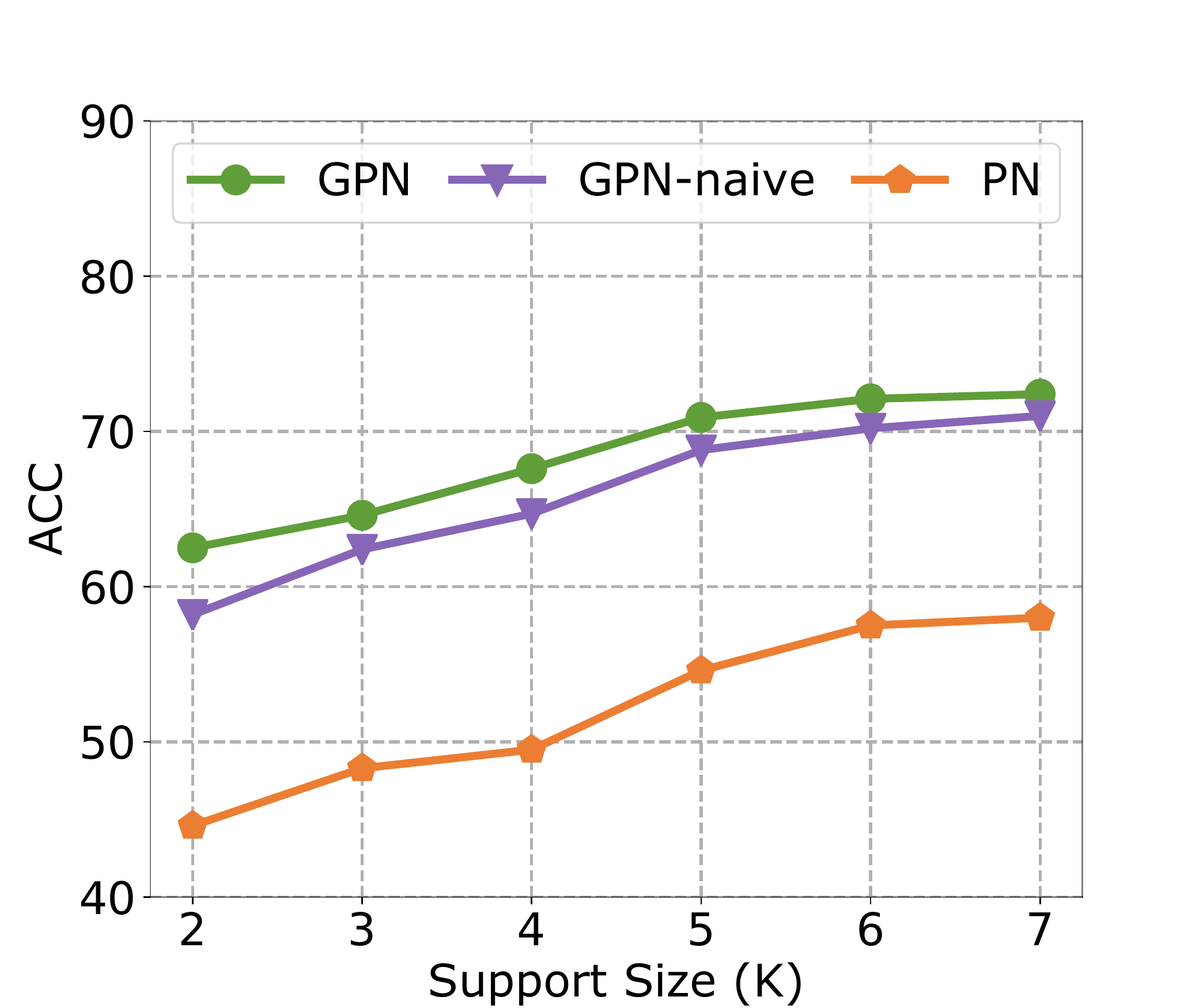}
    }
    \hspace{-0.2cm}
    \subfigure[\textbf{DBLP}] 
    {
    \includegraphics[width=0.23\textwidth]{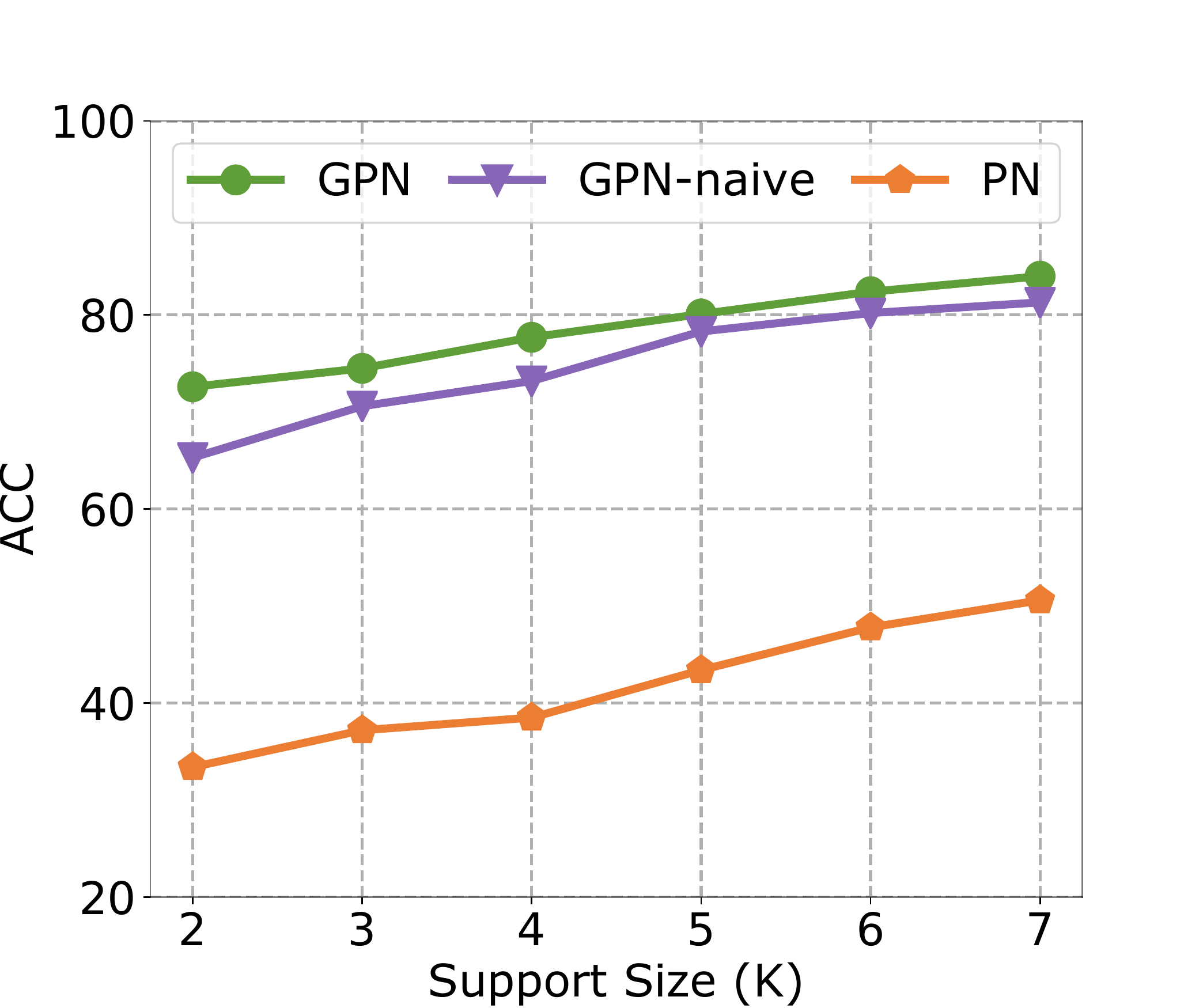}
    }
    \hspace{-0.2cm}
    \subfigure[\textbf{Reddit}]
    {
    \includegraphics[width=0.23\textwidth]{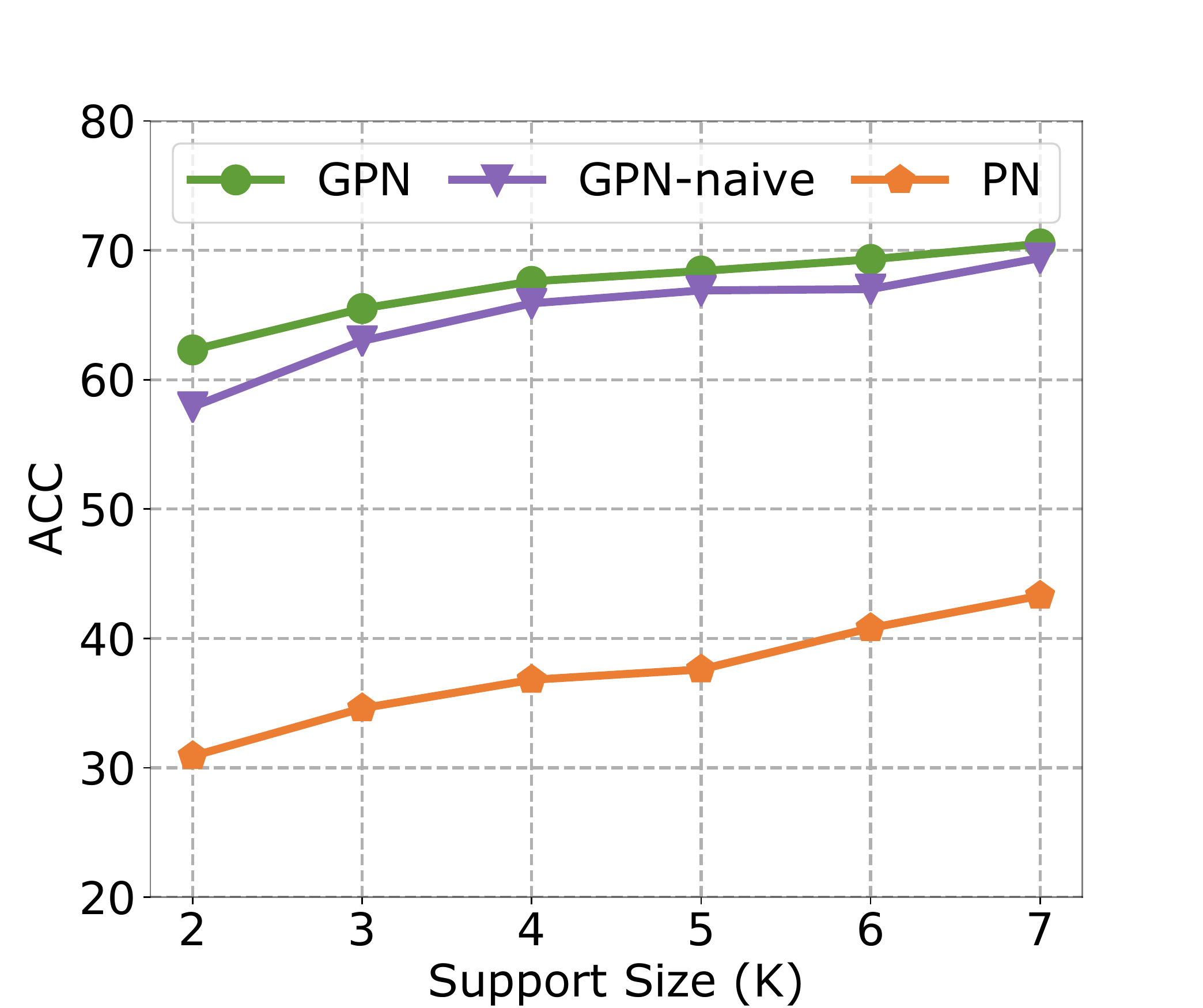}
    }}
    \caption{Performance comparisons \textit{w.r.t.} different support size (5-way $K$-shot).}%

    \label{fig:K}
\end{figure*} 


\subsection{Parameter Analysis \& Ablation Study}
In this section, we conduct extensive experiments to analyze the sensitivity of GPN to the number of node classes ($N$-way), size of the support set ($K$-shot), and query set size. To better understand the contribution of each component, we also include another two methods GPN-naive and PN for \textbf{ablation study}. Note that GPN-naive is a variant of GPN that excludes the \textit{node valuator}, and PN can be considered as a variant of GPN that excludes
the \textit{node valuator} and uses a MLP-based encoder.

\smallskip
\noindent\textbf{Effect of Class Size ($N$-way).}
We first analyze the effect of the test class size, which is controlled by the parameter $N$. Here we keep the 
shot number as $5$, and report the performance changes of the three models by setting different values of $N$. Results on four datasets in terms of Accuracy (ACC) are presented in Figure \ref{fig:N}. From a comprehensive view, the performance of different models decreases as the test class size increases, which is in accordance with our expectation. The main reason is that a larger number of test classes results in a wider variety of node classes to be predicted, which increases the difficulty of few-shot node classification. The performance of PN largely falls behind GPN and GPN-naive since it cannot capture the node dependency information without the GNN-based \textit{network encoder}. In addition to that, we can further observe that GPN consistently outperforms GPN-naive, and the performance margin increases when $N$ becomes larger. It illustrates that the proposed framework GPN is more robust to the number of test classes, which validates the effectiveness of the \textit{node valuator} in GPN for learning more representative class prototypes.

 \begin{figure}[b]
 \vspace{-0.5cm}
    \graphicspath{{figures/}}
    \centering
    \subfigure[\textbf{Amazon-Clothing}] 
    {
    \includegraphics[width=0.485\columnwidth]{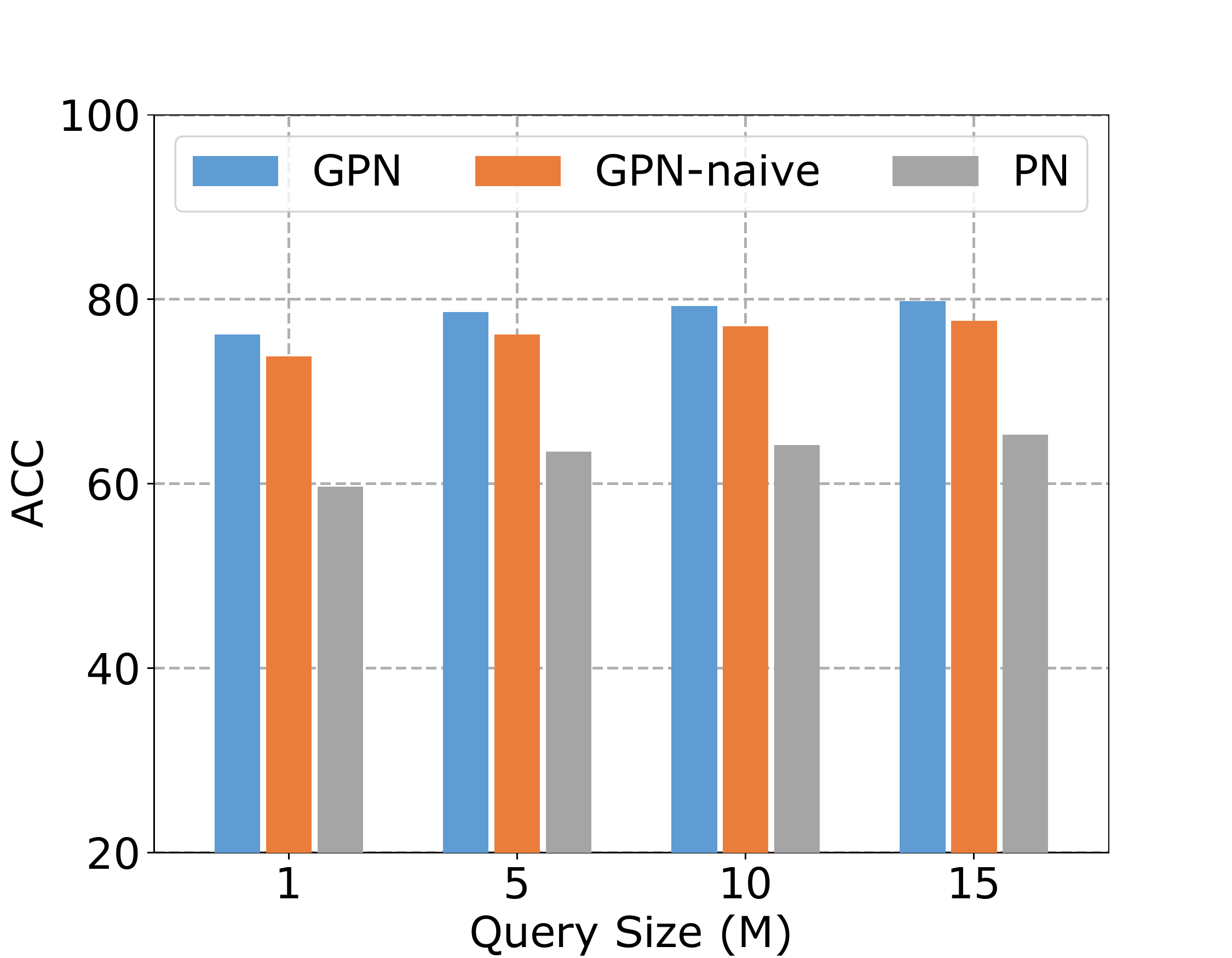}
    }
    \hspace{-0.5cm}
    \subfigure[\textbf{Amazon-Electronics}]
    {
    \includegraphics[width=0.485\columnwidth]{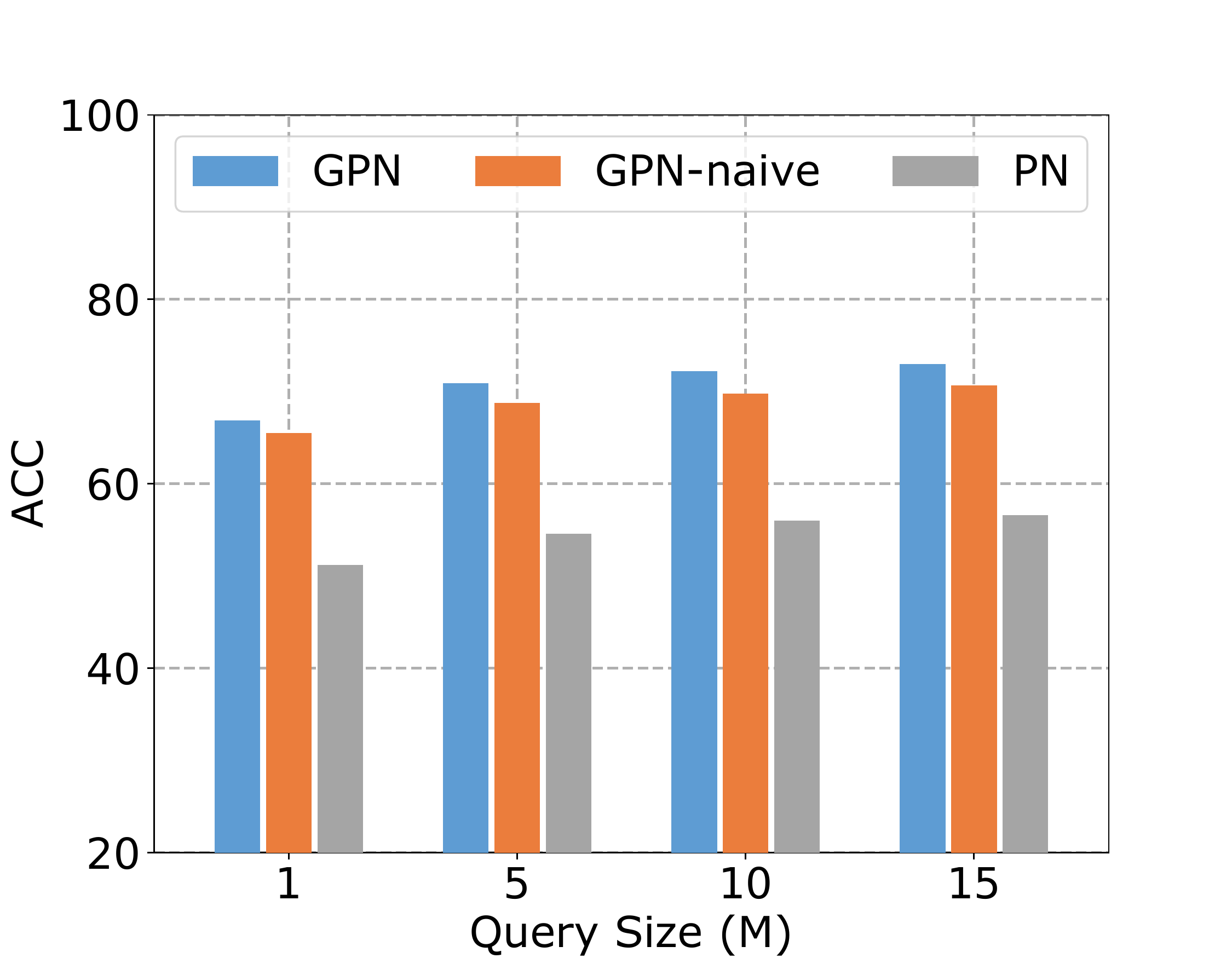}
    }
    \quad
    \subfigure[\textbf{DBLP}] 
    {
    \includegraphics[width=0.485\columnwidth]{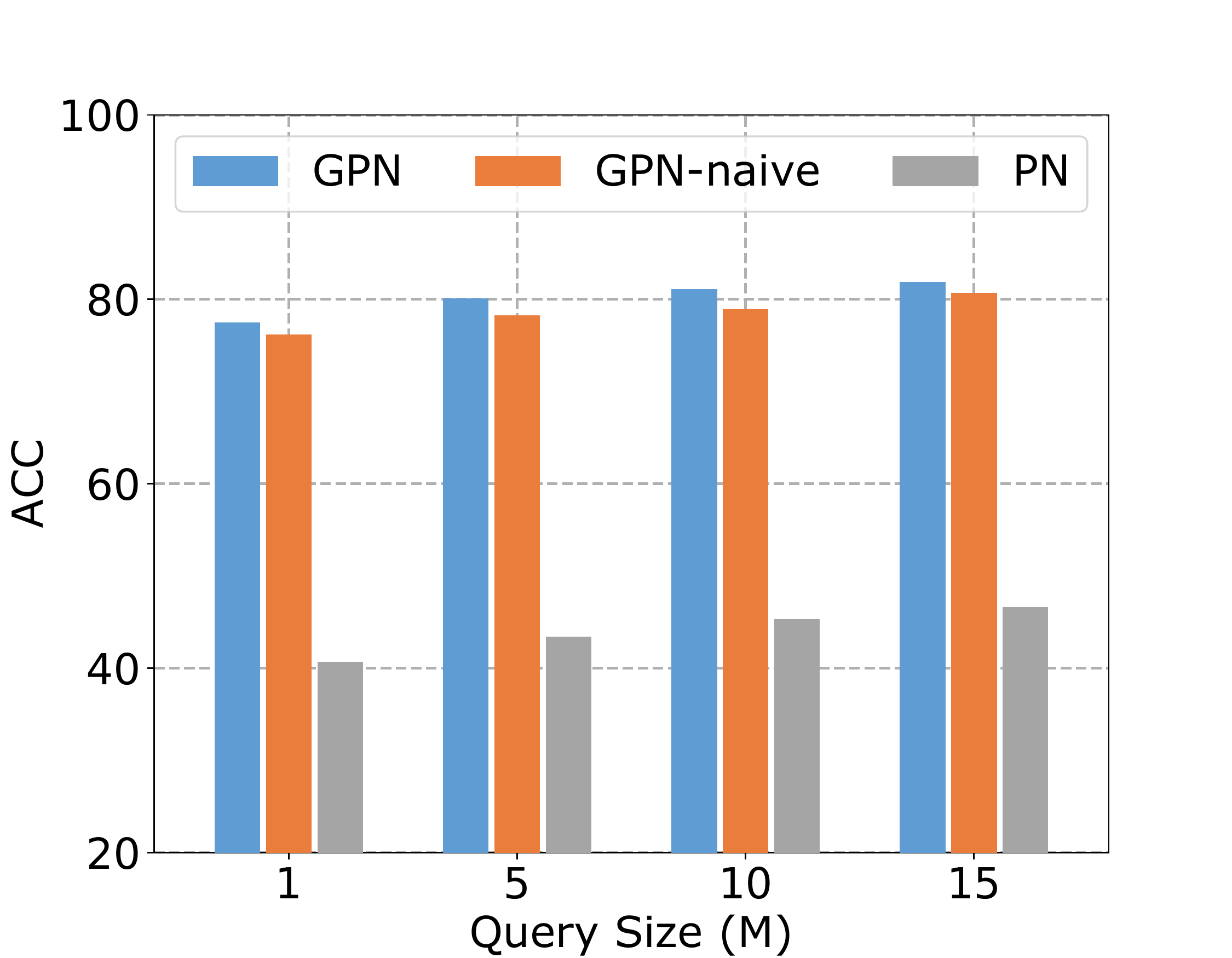}
    }
    \hspace{-0.5cm}
    \subfigure[\textbf{Reddit}]
    {
    \includegraphics[width=0.485\columnwidth]{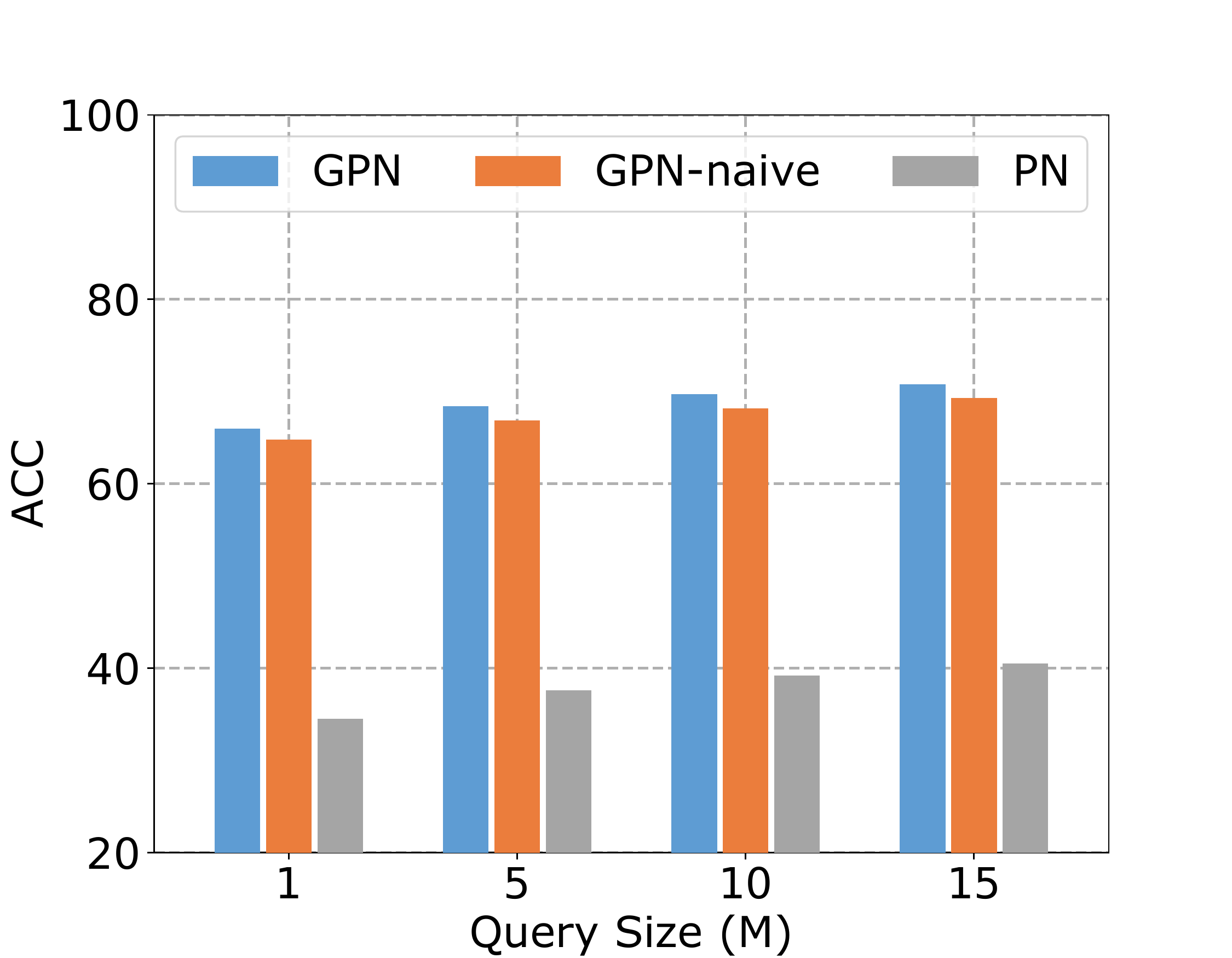}
    }
    \caption{Performance comparisons \textit{w.r.t.} query size (M) on 5-way 5-shot node classification task.}%

    \label{fig:M}
\end{figure}

\smallskip
\noindent\textbf{Effect of Support Size ($K$-shot).}
Next, we investigate the effect of the support size, which is represented by the shot number $K$. By changing the value of shot number $K$ and setting way number $N$ to $5$, we are able to get different model performance. For each dataset, we report the results in terms of Accuracy (ACC) in Figure \ref{fig:K}. From the figure, we can clearly observe that the performance of all the models
increase with the growth of $K$, indicating that larger support set can produce
better prototypes for few-shot classification. PN is unable to achieve satisfactory results due to the inability of modeling attributed networks. More remarkably, we observe that GPN is able to achieve larger improvements over GPN-naive  when the support set size is small. One potential reason could be that GPN-naive is sensitive to noisy data, as it calculates prototype by averaging values over samples with equal weights. Thus, more data is expected to derive reliable prototypes. On the contrary, by estimating the informativeness of each labeled sample, GPN becomes more robust on noisy data and achieves better performance for few-shot node classification.

\begin{figure}[t]
    \graphicspath{{figures/}}
    \centering
    \subfigure[\textbf{Meta-GNN}] 
    {
    \includegraphics[width=0.485\columnwidth]{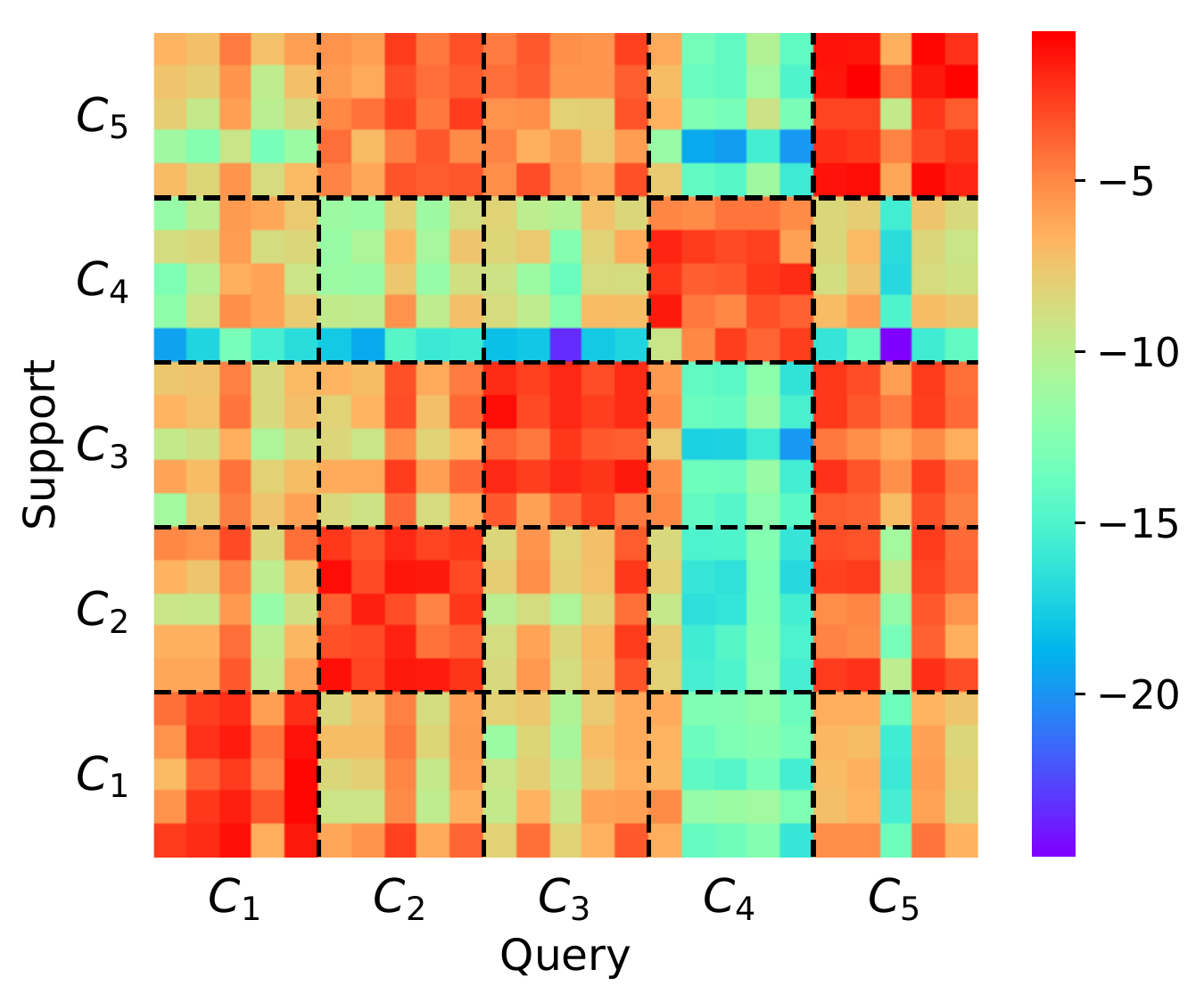}
    }
    \hspace{-0.3cm}
    \subfigure[\textbf{GPN}]
    {
    \includegraphics[width=0.485\columnwidth]{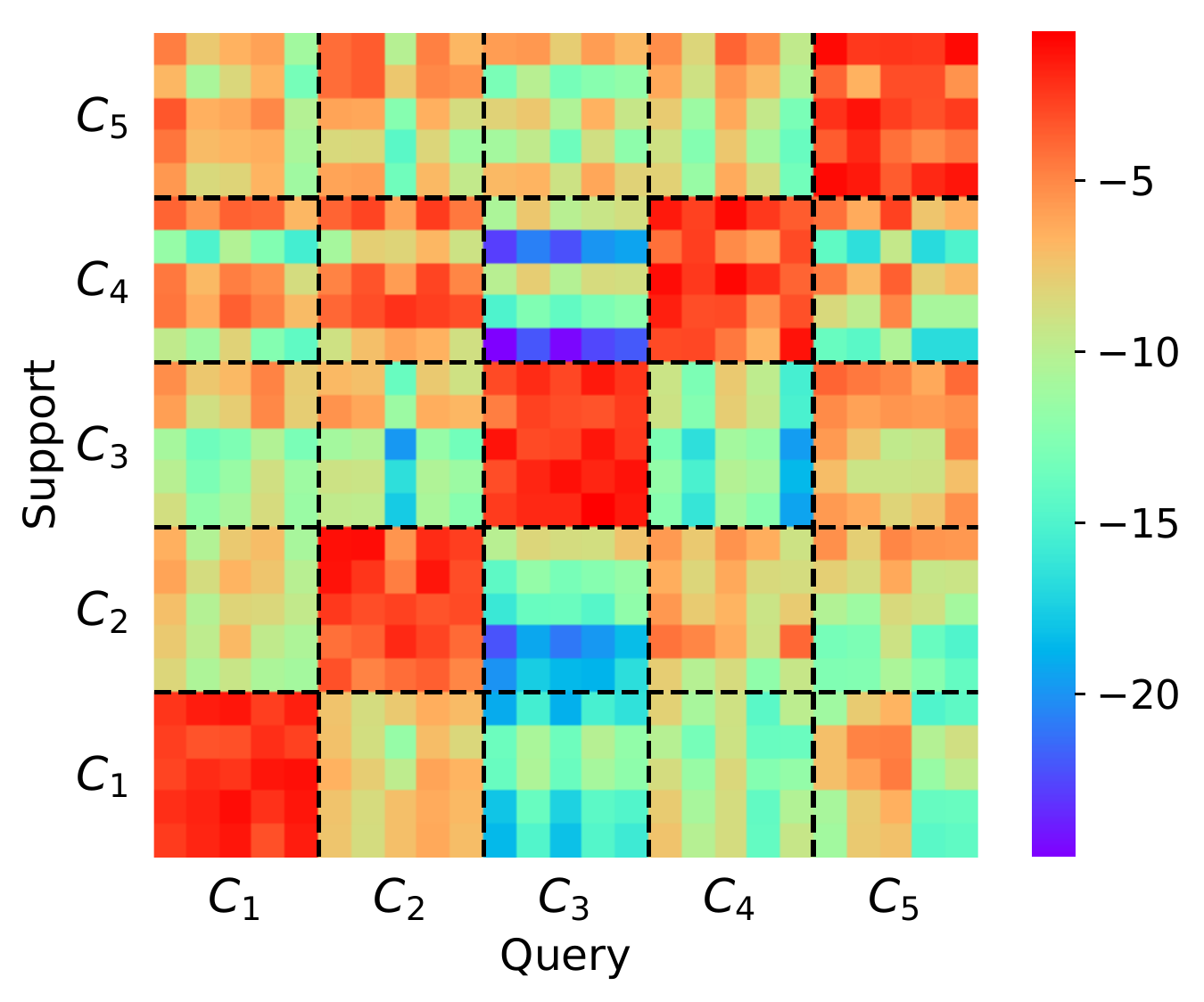}
    }
    \caption{Similarity matrix on DBLP dataset ($5$-way $5$-shot).}
    \label{fig:heatmap}
\end{figure}

\smallskip
\noindent\textbf{Effect of Query Size ($M$-query).}
Although it is a consensus to remain consistent between training and test phase in standard few-shot learning, previous research~\cite{zhang2019variational} claims that not every framework benefits the most from this identical setting. Hence, we further examine the influence of using different query size during training. Here we use the $5$-way $5$-shot task as an example, then change the number of query nodes from each class and report the corresponding results in Figure \ref{fig:M}.
From the reported results, we can observe that increasing query size during training can achieve performance gain on all of the four datasets. One reasonable explanation is that a few-shot learning model can better adapt the knowledge from meta-training tasks with larger query set and further obtain better generalization ability on the target task.

\subsection{Case Study}
Figure \ref{fig:heatmap} shows the similarity matrix learned by the best performing baseline Meta-GNN and our approach on the DBLP dataset, with the same network encoder in a $5$-way $5$-shot task. Here we use the negative Euclidean distance as the similarity metric. Specifically, each cell consists of $5 \times 5$ grids illustrating the divergence between two classes, as well as the intra-class similarities.
To better visualize the results, for GPN, we use the weighted embedding of each support node instead of computing the class prototype. From the figure, we can observe that GPN can better capture the similarities between the support nodes and query nodes from a same class, which validates the robustness and effectiveness of our approach.

%% file: Conclusion.tex
In this paper, we introduce a novel framework Graph Prototypical Networks (GPN) to solve the problem of few-shot node classification on attributed networks. 
Specifically, GPN first extracts node representations via multi-layered graph neural networks considering both node attributes and topological structure. Concurrently, another GNN-based component estimates the informativeness of each labeled node. By integrating those two information modalities, GPN is able to learn highly representative class prototypes in a transferable metric space. Then the label of each query node can be computed by measuring its similarity with prototypes. Moreover, by learning over diverse semi-supervised node classification tasks which can mimic the real test environment in a large number of episodes, GPN can be effectively generalized to the target few-shot classification task. 
The empirical results over four real-world datasets
demonstrate the effectiveness of our proposed model versus the baseline methods in few-shot node classification.

%% file: Appendix.tex
\section{Appendix}
\subsection{Data Accessing and Preprocessing}

All of the graphs in our experiments are constructed from public data sources, whose links are summarized in Table \ref{table:appendix_dataset}. In the following, we provide details on the construction of each graph. 

\smallskip
\noindent\textbf{Amazon-Clothing.} This is a public product dataset ~\cite{mcauley2015inferring} containing the metadata of products in Amazon, ranging from May 1996 to July 2014. The dataset is truncated based on the top-level product category ``Clothing, Shoes and Jewelry''.  Both the product descriptions and substitutable relationships (``also viewed'') between products are included in the metadata. In addition, each product corresponds to a low-level category, e.g., Monopods, LED TVs and DVD Recorders. In this case, each product is denoted as a node and its low-level category is the node label. We select the classes with 100 to 1000 nodes for evaluation and remove those isolated products. Bag-of-words model is applied on product description to obtain the attributes of each node.

\noindent\textbf{Amazon-Electronics.} This dataset is constructed with the products under the category ``Electronics'' in Amazon. Based on the metadata, we use the complementary relationships (``bought together'') between products to create the links. Similar to the previous dataset, the low-level category (e.g., Sunglasses, Garment Bags and Athletic Socks) of each product is used to decide its label. We select the classes with 100 to 1000 nodes for evaluation and omit those isolated products.


\noindent\textbf{DBLP.} We use the public DBLP dataset (version v11) \cite{tang2008arnetminer} which covers information (e.g., abstract, authors, references and venue) for all the papers available on DBLP before May 2019. In the experiment, we focus on venues which have been lasting for at least 20 years and published 100 to 1000 papers. All the isolated nodes with no link are excluded. Then we apply the bag-of-word model on the abstract of each node to generate the node attributes. 

\noindent\textbf{Reddit.} To construct this post-to-post graph, we use the public dataset \cite{hamilton2017inductive} sampled from Reddit with all the posts made in September 2014. Each post belongs to one of the 50 large communities in Reddit. With the off-the-shelf 300-dimensional GloVe CommonCrawl word vectors \cite{pennington2014glove}, for each post, we concatenate the emebedding of post title, the average embedding of all the comments, the post's score and the number of comments.


\begin{table}[!h]
\centering
\caption{Links for accessing the original datasets}
\resizebox{0.48\textwidth}{!}{%
\begin{tabular}{ll}
\toprule
 \textbf{Datasets} & \multicolumn{1}{c}{\textbf{Links}} \\ \hline
Amazon-Clothing & \begin{tabular}[c]{@{}l@{}}http://snap.stanford.edu/data/amazon/productGraph/ \\ categoryFiles/meta\_Clothing\_Shoes\_and\_Jewelry.json.gz\end{tabular} \\ \hline
Amazon-Electronics & \begin{tabular}[c]{@{}l@{}}http://snap.stanford.edu/data/amazon/productGraph/ \\ categoryFiles/meta\_Electronics.json.gz\end{tabular} \\ \hline
DBLP & \url{https://www.aminer.cn/citation} \\ \hline
Reddit & \url{http://snap.stanford.edu/graphsage/\#datasets} \\ 
\bottomrule
\end{tabular}
}
\label{table:appendix_dataset}
\end{table}